\documentclass[authoryear,preprint,12pt]{elsarticle}

\usepackage{graphicx}

\usepackage{amssymb,amsthm}
\usepackage{graphicx} 
\usepackage{epsfig} 
\usepackage{subcaption}
\usepackage{amsmath,amsfonts} 
\usepackage{bm}
\usepackage{float}
\usepackage[margin=0.85in]{geometry}

\usepackage{psfrag}

\usepackage{hyperref}
\hypersetup{
     colorlinks   = true,
     citecolor    = gray
}

\renewcommand*{\contentsline}[3]{\csname l@#1\endcsname{#2}{}}
\usepackage{tocloft}

\usepackage{color}

\usepackage{enumitem}

\usepackage{epsfig} 
\usepackage{amsmath} 
\usepackage{amssymb}  
\usepackage{microtype}
\usepackage{epstopdf}
\usepackage{booktabs}
\usepackage{breqn}
\usepackage{bm}

\usepackage{multirow}
\usepackage[ruled,vlined,linesnumbered]{algorithm2e}

\usepackage{tikz}
\usetikzlibrary{shapes,arrows,chains}
\tikzset{middlearrow/.style={
        decoration={markings,
            mark= at position 0.6 with {\arrow{#1}} ,
        },
        postaction={decorate}
    }
}

\usepackage{multirow}
\usepackage{url}
\usepackage{makecell}
\usepackage{relsize}
\usepackage{wrapfig}
\usepackage{lscape}
\usepackage{rotating}

\usepackage{mathtools}
\DeclareMathAlphabet{\pazocal}{OMS}{zplm}{m}{n}
\makeatletter
\newcommand{\vast}{\bBigg@{3}}
\makeatother
\makeatletter
\renewcommand*\env@matrix[1][\arraystretch]{%
  \edef\arraystretch{#1}%
  \hskip -\arraycolsep
  \let\@ifnextchar\new@ifnextchar
  \array{*\c@MaxMatrixCols c}}
\makeatother

\newcommand{\Tau}{\mathrm{T}}
\journal{Transportation Research Part C}

\begin{document}
\begin{frontmatter}

\title{Dynamic optimal congestion pricing in multi-region urban networks by application of a Multi-Layer-Neural network}

\author{Alexander Genser\corref{cor1}}
\ead{gensera@ethz.ch}

\author{Anastasios Kouvelas}
\ead{kouvelas@ethz.ch}

\cortext[cor1]{Corresponding author. Tel.:  +41-44-632-75-19.}

\address{Institute for Transport Planning and Systems, Department of Civil, Environmental and\\Geomatic Engineering, ETH Zurich, CH-8093 Zurich, Switzerland}


\begin{abstract}
Traffic management by applying congestion pricing is a measure for mitigating congestion in protected city corridors. As a promising tool, pricing improves the level of service in a network and reduces travel delays. However, real-world implementations are restricted to static pricing, i.e., the price is fixed and not responsive to the prevailing regional traffic conditions. Dynamic pricing overcomes these limitations but also affects the user’s route choices. This work uses dynamic pricing's influence and predicts pricing functions to aim for a system optimal traffic distribution. The framework models a large-scale network where every region is considered homogeneous, allowing for the Macroscopic Fundamental Diagram (MFD) application. We compute Dynamic System Optimum (DSO) and a Quasi Dynamic User Equilibrium (QDUE) of the macroscopic model by formulating a linear optimization problem and utilizing the Dijkstra algorithm and a Multinomial Logit model (MNL), respectively. The equilibria allow us to find an optimal pricing methodology by training Multi-Layer-Neural (MLN) network models. We test our framework on a case study in Zurich, Switzerland, and showcase that (a) our neural network model learns the complex user behavior and (b) allows predicting optimal pricing functions. Results show a significant performance improvement when operating a transportation network in the DSO and highlight how dynamic pricing influences the user's route choice behavior towards the system optimal equilibrium.   
\end{abstract}


\end{frontmatter}



\section{INTRODUCTION}
\label{S:1-Intro}
The fact that more and more people live in cities puts significant pressure on the mobility services of urban areas. One major challenge of today's transportation systems is the mitigation of congestion. Therefore, the traffic management domain has proposed various technologies to tackle rising traffic demand in the last decades. Research has shown several effective microscopic approaches, such as, e.g., optimal traffic light control and macroscopic methods, where Perimeter Control (PC) is well recognized. PC allows a significant reduction of user delay in a protected region by controlling traffic lights at the region border~\citep{ref:perimeter_c1, ref:perimeter_c2, kouvelas2017enhancing}.
Nevertheless, methods, such as PC, do not consider external effects when focusing on car traffic. Air pollution, noise, accidents, congestion, and space occupation are examples of costs the road users do not have to accommodate for. Hence, this results in adverse effects on the performance of a traffic system, the environment, and the economy \citep{ref:hansen}.

\cite{ref:pigou1920} showed early on in his extensive work on external costs that congestion pricing (often also denoted as road pricing) is a promising tool. Practical evaluations of field implementations in, e.g., London, Stockholm, and Singapore underline the effectiveness of congestion pricing~\citep{ref:eliasson}. The impact of rising traffic demand is mitigated, internalization of external effects and a reduction in performance metrics such as travel times, vehicle kilometers traveled, or travel delays is achieved.

Although there is a broad agreement on the efficiency of congestion pricing, the design of the pricing scheme itself is still tackled by several research domains. As already reviewed by~\cite{ref:Lindsey2000_Traffic_Cong} the fundamental work from~\cite{ref:vickrey} was one of the first that shows the potential of congestion pricing to influence the travel behavior (i.e., the route and mode choice). Also, the work claims the need to set prices reflecting the current traffic state in a network; i.e., if a city experiences congestion, tolls need to react dynamically with a specific magnitude. Note that prices and tolls are used as synonyms in this work. In recent years, several theoretical (micro- and macroscopic) studies have shown that extending a transportation system with a dynamic pricing scheme can further improve the performance of a transportation network (e.g.,~\cite{, ref:zheng_dynamic_pricing, ref:kachroo, ref:zheng16, ref:Gu2018JDTT, ref:Yang2019_Joint_CP_PC}. 

However, the implementation of congestion pricing on a link-level has been found as not practical. High investments to upgrade the infrastructure and regulation issues (e.g., the infrastructure operator needs to provide an alternative non-tolled route) are faced in practice. One of the first works that try to overcome this challenge by tackling congestion pricing at the macroscopic level was published by~\cite{ref:zheng_dynamic_pricing}. The work utilizes the Macroscopic Fundamental Diagram (MFD) to derive an optimal cordon-based pricing scheme.~\cite{ZHENG2020_area_based_pricing}, ~\cite{GU2021_method_comparision}, and~\cite{CHEN2021_travel_cost_perception} follow this approach and show the advantages of an aggregated approach on network-level. 

Besides the comprehensive findings of these studies, they lack an analytical and efficient formulation of a real-time system optimum. This is of great interest, as the optimal quantities can be utilized to derive optimal pricing functions. 
In the present work, we focus on a multi-region network model based on~\cite{ref:isik_model_rg} to find the optimal macroscopic pricing scheme with the application of supervised machine learning. The defined urban regions are considered homogeneous with different characteristics (i.e., size, capacity, average trip length) in the heterogeneous traffic network. A well-defined MFD characterizes every region. The determination of Dynamic System Optimum (DSO) is solved by reformulating the nonlinear model into a linear program and applying several approximations based on the work by~\cite{ref:Genser2020TRB} with a Linear Rolling Horizon Optimization (LRHO); i.e., the optimal route choice is determined. Implementation of the Quasi Dynamic User Equilibrium (QDUE) is based on the utilization of Djikstra algorithm to find the shortest paths and a multinomial Logit (MNL) model to determine the user's route choices. To determine the optimal time-varying pricing functions, we train deep neural networks (more specifically, a Multi-Layer-Neural (MLN) network) models that capture the complex user's route choice behavior and predict the generalized trip costs in every region.  

The application of the proposed framework, including the pre-trained pricing prediction models, allows us to highlight the following contributions: (a) 
the formulation of an efficient and linear program to find real-time solutions for the nonlinear optimal route guidance problem (i.e., the DSO); (b) the design of MLN network models that allow a generic application to learn the user's route choice behavior and predict their generalized trip costs; (c) the derivation of demand-specific pricing functions for optimal tolling in a multi-region network.

The remainder of this paper is organized as follows: Section~\ref{sec:related_work} points the reader to related work on the derivation of user and system equilibrium and optimal pricing. Section~\ref{sec:model} introduces the utilized macroscopic simulation model. Section~\ref{sec:methodology} elaborates on the applied methodology. First, the derivation of DSO with all steps to linearize the problem is introduced. The chapter continues with the QDUE by applying Djikstra algorithm and MNL. The optimal toll derivation with a MLN network model is presented at the end of Section~\ref{sec:methodology}. The methodology is applied to a case study in Zurich, Switzerland, with results for DSO, QDUE, and the optimal pricing functions (Section~\ref{sec:casestudy}). The paper closes with a conclusion and future work in Section~\ref{sec:conclusion}.

\section{RELATED WORK} \label{sec:related_work}
The general idea of congestion pricing was picked up already decades ago by~\cite{ref:pigou1920} followed by works such as~\cite{ref:knight, ref:vickrey}. Since then, optimal pricing problems have been tackled for microscopic and macroscopic traffic models. Generally, the literature differentiates between two types of pricing problems: First-best pricing is defined by pricing every link in a network efficiently. Thus, network modeling on the microscopic level is essential and information of every link must be available in real-time. Works such as~\cite{ref:Beckmann_MSCP} suggest deriving tolls for first-best pricing with the concept of marginal social cost pricing. Consequently, the tolls represent an equivalent of the negative externalities caused to other users in the transportation network.  Differently, ~\cite{ref:Bergendorff1997} and~\cite{ref:Hearn_first_best_toll} show that toll vectors exist for fixed demand scenarios and define the first-best toll set based on the concepts of user equilibrium and system optimum. Finally,~\cite{ref:Yildirim_first_best_toll} extend the proposed solutions from~\cite{ref:Bergendorff1997} and~\cite{ref:Hearn_first_best_toll} for demand uncertainties with a General Variable Demand (GVD) model. Although the application of tolls derived by solving the first-best pricing problem guarantees the operation at the system optimum, the practicality has been critical discussed.~\cite{ref:Lindsey2000_Traffic_Cong} argues that even with electronic tolls (users do not have to stop at a toll which causes travel delays), the investment and operational costs are high. Several countries' regulations force the infrastructure operator to provide an alternative non-tolled route. Furthermore, it is unlikely that a toll system is implemented in the whole network at once, which is a constraint for first-best pricing. 

Therefore, recent research focuses on the second-best pricing problem, where only a subset of the links is utilized for pricing. Considering a small toy network, represented as a graph,~\cite{ref:meng2012optimaltoll} and~\cite{ref:chung2012optimaltollDemU} are formulating optimization problems to derive optimal tolls with a bi-level cellular particle swarm optimization and (considering demand uncertainties) a mixed-integer problem, respectively. Both works focus on determining prices by utilizing the total distance traveled on priced arcs in a network. Furthermore, an optimal speed-based pricing design using the average travel speed has been proposed by~\cite{ref:Lui2013SpeedToll}. Finally, a joint model incorporating the travel distance and time (Joint distance and time toll (JDTT)) was introduced by~\cite{ref:Liu2014JDDT}. 
Methodologies of the aforementioned papers are operating at the link level, which remains challenging when one considers a city center corridor with a high number of links (holds for the first and second-best toll problem).

Besides, sophisticated modeling incorporates a dynamic traffic assignment, making the computation of dynamic traffic equilibria (i.e., QDUE and DSO) relatively expensive. Nevertheless, this is an essential procedure to evaluate the applied dynamic pricing scheme. 
~\cite{ref:vanEssen} collects a literature review on the role of travel information and stresses the importance to push a system from the user equilibrium (i.e., people are behaving selfishly in their route choice and try to maximize their benefit) to a more efficient system optimum (i.e., a part of the network users need to act in a non-selfish way by choosing an alternative route which might result in, e.g., a longer travel time). The interested reader is referred to works such as~\cite{ref:Amirgholy2017} or~\cite{ref:Zhong2020_DSO} for a detailed derivation of macroscopic traffic model's DSO. One way to direct a transportation network towards the system optimum is by influencing the user's route choice with travel information. Different systems have been utilized in the past to provide users with information about the current toll to enter a protected region (e.g., a website or mobile app that provides the current price or information systems on the highway displaying the current toll one would need to pay)~\citep{ref:Suihi}. With the advancement of vehicle technology, toll data can even be provided in real time to the user. Consequently, congestion pricing can not only be utilized as a general solution for reducing car traffic demand (i.e., the user's mode choice or departure time is influenced) or the internalization of external effects. The user's route choice can be affected, leading to a better distribution of traffic in the network and, consequently, better system performance.

Accounting for given limitations of microscopic modeling of congestion pricing, other works have focused on a macroscopic approach using multi-region models and MFD. To the authors' best knowledge, one of the first works considering MFD to obtain optimal pricing was published by~\cite{ref:zheng_dynamic_pricing}. Utilizing an agent-based simulator, the concept of MFD, and a proportional controller, dynamic cordon-based pricing is shown as an efficient tool to save travel times and ease congestion in the cordon.~\cite{ref:Simoni2015MarginalCP} design an area-based pricing scheme with the concept of marginal costs. The derived pricing methodology is applied to the simulator MATSim.
Nevertheless, the proposed method does not include a feedback strategy.~\cite{ref:Gu2018JDTT} investigate several pricing methodologies with simulation-based optimization and feedback control. The work combines a microscopic simulator with a Proportional-Integral (PI) control utilizing the MFD. This approach allows maintaining a protected region at the critical vehicle accumulation (corresponding to the maximum vehicle flow) and calculating prices based on the link-based distance and time traveled. Nevertheless, the iterative approach introduces a heavy dependency on a simulator to derive the MFD and the link-based prices; moreover, no comparison to traffic equilibria is performed. 
~\cite{ZHENG2020_area_based_pricing} propose another area-based pricing system by considering heterogeneous user groups; i.e., the user groups are characterized with different Value of Times (VOT). Then, by applying a network aggregated two-region model and the concept of MFD, the paper derives fair tolls. Also, with MFD,~\cite{CHEN2021_travel_cost_perception} propose another PI control to optimize tolls by utilizing MATsim as a simulator. Besides, the Cumulative Prospect Theory (CPT) is applied to (a) reduce the peak-hour demand with tolling and (b) model the level of service experienced by network users. 

As the derivation of pricing for large-scale urban networks requires the consideration of complex relationships (e.g., network structure, traffic flow theory, route choice, etc.), several recently published works tackle the problem with machine learning models. E.g.,~\cite{ref:DL_MOHANTY2020} forecast traffic congestion within a region by applying a Long-Short-Term-Memory (LSTM) neural network and derive a neighborhood congestion score. The work shows that the LSTM model is useful for the optimal pricing problem.~\cite{ref:DL_Shukla2020} also utilized a spatially-induced LSTM to predict the current traffic conditions. The output of their LSTM model and road network-related parameters are then fed to a proposed algorithm that allows the determination of tolls. Apart from neural networks also reinforcement learning has gained rising attention to tackle optimal pricing with promising results; the interested reader is referred to works such as,~\cite{ref:RL_SATO2021, ref:RL_Zhu, ref:RL_Mirzaei2018}. 

\section{MACROSCOPIC MULTI-REGION MODELING} \label{sec:model}
\begin{table}[!b]
\centering
\begin{tabular}{lcl}
\toprule
Notation & Unit & Description \\
\midrule
     $t$        &  [s]  & Continuous simulation time step \\
     $T$        & [s] & Simulation time \\
     $\mathcal{R}$     &    -  &  Set of regions\\
     $I$        &  -    & Origin region and element of $\mathcal{R}$  \\
     $J$        &  -    & Destination region and element of $\mathcal{R}$  \\
     $\mathcal{N}_I$ & - & Set of all neighboring regions of $I$ \\
     $H$ & - & Stop-over region and element of $\mathcal{N}_I$ \\
     $K$    &   -   &  Total number of regions \\
     $N_{II}, N_{IJ}$ & [veh] & Vehicle accumulation from origin $I$ to $I$ or $J$, respectively \\
     $N_I, N_H$ & [veh] & Aggregated vehicle accumulation for region $I$ and $H$ \\
     $G(\cdot)$ & [veh/s] &  Outflow MFD of a region $I$ \\
     $\bar{L}_I$ & [m] & Average trip length in region $I$ \\
     $Q_{II}, Q_{IJ}$ & [veh/s] & Traffic demand from $I$ to $I$ or $J$, respectively \\
     $M_{II}$ & [veh/s] & Internal flow from region $I$ to $I$ \\
     $M_{IHJ}$ & [veh/s] & Transfer flow from origin $I$ over $H$ to destination $J$ \\
     $\theta_{IHJ}$ & - & Route choice variable (QDUE) from region $I$ to  $J$ via $H \in \mathcal{N}_I$ \\
     $n$ & - & Polynomial degree for MFD design \\
     $a, b, c$ & - & Polynomial coefficients for MFD design \\
     $\tilde{M}_{IHJ}$ & [veh/s] & Transfer flow preventing a region from overflow \\
     $C(\cdot)$ & [veh/s] & Capacity function \\
     $t_r, t_f, t_c$ & [s] & Demand parameters for rising, falling and const. magnitude time. \\ 
     $Q_t$ & [veh/s] & Demand magnitude for $Q_{II}$ and $Q_{IJ}$\\
     \bottomrule
\end{tabular}
\caption{Notation for the multi-region model. The time component $t$ is omitted for the reader's convenience.}
\label{tab:notation_model}
\end{table}
In this paper, the traffic network is modeled as a multi-region network partitioned into homogeneous regions. The notation of used variables denotes Table~\ref{tab:notation_model}. The homogeneous regions are defined by $\mathcal{R} = \{1, 2, \ldots, K\}$, where $K$ is the total number of regions. Every region from $\mathcal{R}$ is modeled with a well-defined MFD, represented by the function $G(N_I(t))$. $N_I(t)$ denotes the aggregated vehicle accumulation of a region $I$ at time $t$. Consequently, the dynamic equations can be defined in continuous time as follows:
\begin{equation}
	\frac{dN_{II}(t)}{dt} = Q_{II}(t) - M_{II}(t) + \sum_{H \in \mathcal{N}_I} M_{HII}(t),
\label{eq:dyn1}
\end{equation}
\begin{equation}
	\frac{dN_{IJ}(t)}{dt} = Q_{IJ}(t) - \sum_{H \in \mathcal{N}_I} M_{IHJ}(t) + \sum_{H \in \mathcal{N}_I; H \neq J} M_{HIJ}(t),
\label{eq:dyn2}
\end{equation}
where indices $I \in \mathcal{R}$, $H \in \mathcal{N}_I$ and $J \in \mathcal{R}$ represent the origin, stop-over, and destination region, respectively. Variables $N_{II}(t)$ and $N_{IJ}(t)$ denote vehicle accumulations of region $I$ that have final destination region $I$ and $J$, respectively. $\mathcal{N}_I$ is a set containing all neighboring regions of $I$. Internal demand within one region is defined by $Q_{II}(t)$; moreover, demands with origin $I$ and destination $J$ are denoted by $Q_{IJ}(t)$. Note that $Q_{II}(t)$ and $Q_{IJ}(t)$ are exogenous signals. Intra- and inter-regional flows are computed by functions $M_{II}(t)$ and $M_{IHJ}(t)$ representing internal flows in a region and transfer flows from region $I$ to $H$ (with final destination $J$), respectively, defined as follows:
\begin{equation}
	M_{II}(t) = \frac{N_{II}(t)}{N_{I}(t)}G(N_I(t)),
\label{eq:dyn3}
\end{equation}
\begin{equation}
	M_{IHJ}(t) = \theta_{IHJ}(t)\frac{N_{IJ}(t)}{N_{I}(t)}G(N_I(t)).
\label{eq:dyn4}
\end{equation}
Variables $\theta_{IHJ}(t)$ represent the route choices at time $t$; for their computation, an implementation of Dijkstra shortest path algorithm in combination with a MNL model is utilized for the computation of QDUE. To find the optimal route guidance (i.e., DSO splitting rates), a linear optimization problem is solved (see Section~\ref{sec:eq_derivation} for derivations).

\begin{figure}[!b]
    \centering
    \includegraphics[width=0.5\textwidth]{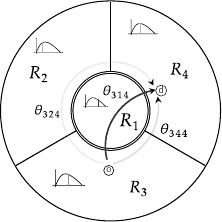}
    \caption{Four-region model with potential routes from origin $I=3$ to destination $J=4$.}
    \label{fig:model_routes}
\end{figure}

The sequence of regions a user can traverse in the proposed model is not arbitrary. If the indices $IHJ$ are parametrized with $I=J$ (e.g.,\ $IHJ = 131$), paths are restricted. This assumption does not allow for unrealistic path choices and improves the quality of the model. An example of a four-region model and the allowed possibilities to move from an origin (o) to a destination (d) are shown in Figure~\ref{fig:model_routes}. The black route here represents the user's choice traversing from $I=3$ via $H=1$ to the destination $J=4$. Hence, this represents one potential solution of the DUE, although it might be more beneficial for the whole four-region system that the user travels a path where the splitting rates $\theta_{324}$ or $\theta_{344}$ apply. 

Note that the transfer flows need to be restricted by~(\ref{eq:dyn5}). The minimum among incoming transfer flow or maximum region capacity is considered, preventing a region from accepting incoming flows that exceed capacity (overflow, $	\tilde{M}_{IHJ}(t)$). The latter is modeled with function $C_{IHJ}(N_H(t))$ (the reader is referred to~\cite{ref:isik_model_rg} for the modeling of function $C(\cdot)$).  
\begin{equation}
		\tilde{M}_{IHJ}(t) = \min\big(C_{IHJ}(N_H(t)), \theta_{IHJ}(t)\frac{N_{IJ}(t)}{N_{I}(t)}G(N_I(t))  \big).
	\label{eq:dyn5}
\end{equation}

Elements of set $\mathcal{R}$ are considered as homogeneous and can, therefore, be characterized by a well-defined MFD. Previous works are using mathematical relationships to model an MFD represented as a polynomial of degree $n$ (e.g.,\ in~\cite{ref:Geroliminis08}. Furthermore, other approximations, such as an exponential function or a novel method proposed by~\cite{ref:ambuehl} proposing to estimate the MFD from measurement data, are applied. However,  several methods suffer from function parameters that lack physical meaning and might introduce problems with optimization procedures. Hence, the current work models function $G(\cdot)$ with a polynomial of degree $n=3$. With commonly utilized mathematical procedures a polynomial is easy to fit and is simpler to handle when linearizing an optimization problem (concavity and continuity). $G(\cdot)$ is defined as follows:  

\begin{equation}
    G(N_I) = \Big (aN^3_I(t) + bN^2_I(t) + cN_I(t)\Big)/\bar{L_I}.
\end{equation}

Function $G(\cdot)$ is the estimated outflow [veh/s] with respect to $N_I$; the coefficients $a$, $b$, $c$ are derived by fitting the polynomial to e.g., derived measurement data from loop detectors. The variable $\bar{L}_I$ denotes the average trip length. 

\begin{figure}[!b]
    \centering
    \includegraphics[width=1.0\textwidth]{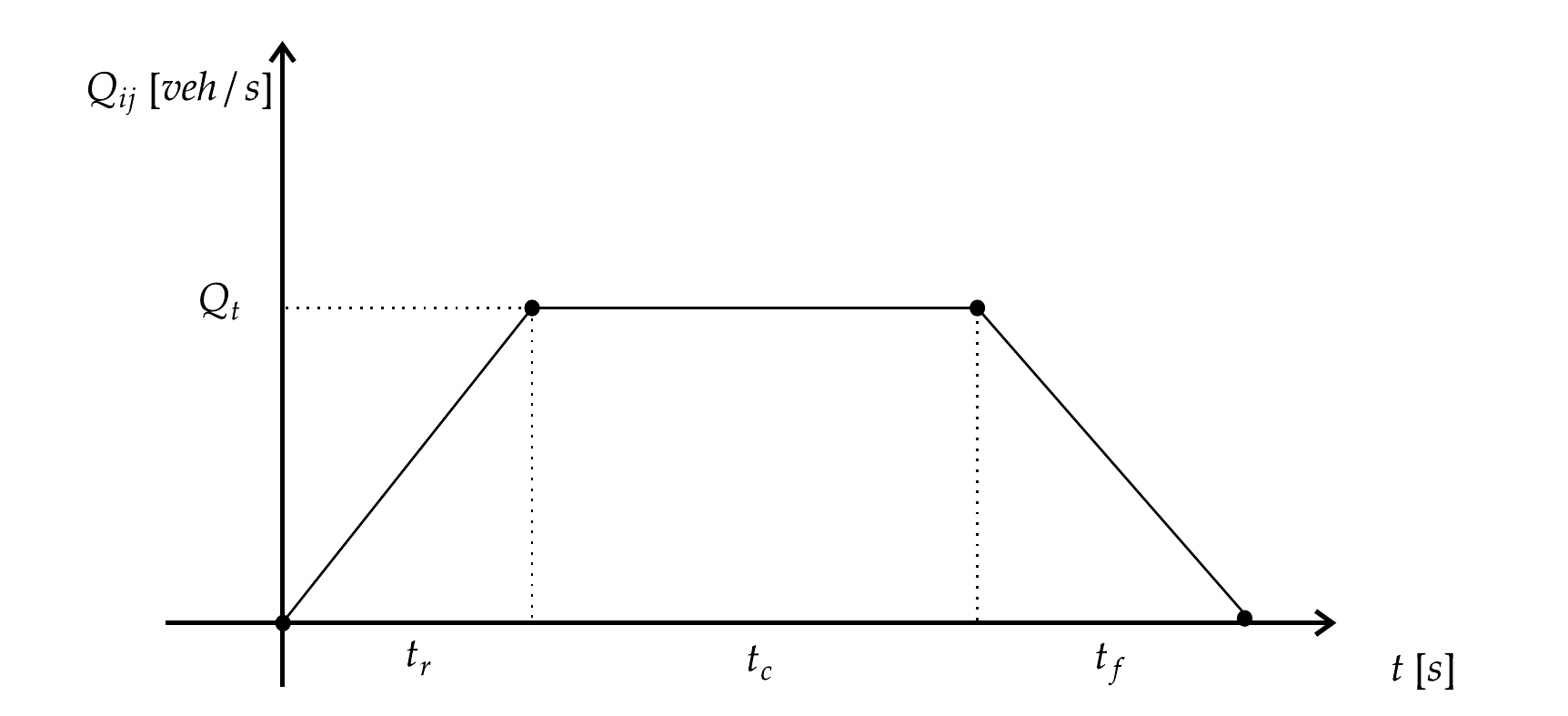}
    \caption{Representation of demand patterns as trapezoids for $Q_{IJ}$.}
    \label{fig:demand}
\end{figure}

To model a realistic demand-supply system, the simulation plant receives demand patterns as trapezoids. A trapezoid is defined as an euclidean geometry shape by specifying the rising time $t_{r}$ [s], falling time $t_{f}$ [s], time that the demand remains at constant magnitude $t_{c}$ [s], and demand magnitude $Q_t$ in [veh/sec]. Figure~\ref{fig:demand} shows a graphical representation of the parameter definition. 
Often these parameters are found by generating random numbers that satisfy the given application requirements. In the current work, an optimization procedure from~\cite{ref:kosmatopoulos} is utilized to find appropriate parameters $t_{r}$, $t_{f}$, $t_{c}$, and $Q_t$, producing a desired simulation scenario (e.g.,\ two congested and two uncongested regions). By setting a target accumulation per region on the MFD curves, different scenarios for testing the optimal route guidance determination and pricing methodology can be generated efficiently for a simulation time of $T$~\citep{ref:GenserSTRC19}. 

The next section of this work presents the utilization of the presented simulation model and all required methodology parts to derive the proposed optimal congestion pricing method. 

\section{METHODOLOGY}
\label{sec:methodology}
\begin{figure}[!b]
    \centering
    \includegraphics[width=0.9\textwidth]{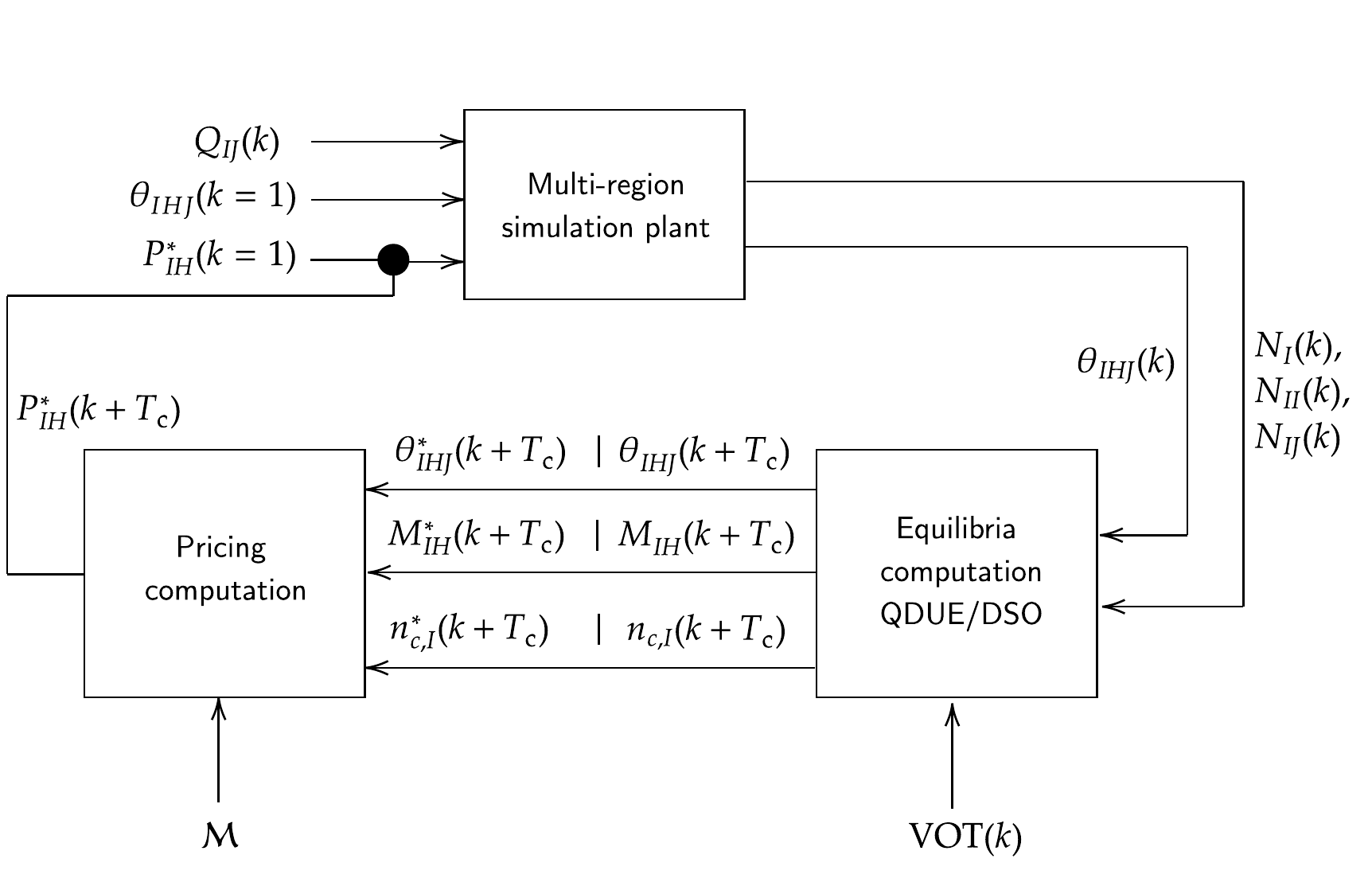}
    \caption{Block diagram of the optimal congestion pricing methodology.}
    \label{fig:block_diagram_method}
\end{figure}

Utilizing the simulation model from Section~\ref{sec:model} we introduce the methodology that allows us to compute (a) the traffic equilibria and (b) predict the generalized costs, and (c) utilize this quantities to derive optimal pricing for the multi-region network. Figure~\ref{fig:block_diagram_method} depicts a block diagram with all the core components that are introduced in this section. We utilize the multi-region simulation plant from Section~\ref{sec:model} in a discretized form to simulate traffic scenarios for a given exogenous demand profile. To compute the equilibria, i.e., DSO and QDUE, a linear formulation of an optimal route guidance optimization problem is used for the DSO. For the QDUE the Dijstrka algorithm (to compute the shortest path) and an MNL (to model the travel behavior) are utilized. Finally, we feed the outputs from the equilibria derivation to the pricing computation. Within this procedure, pre-trained MLN networks are utilized to derive the optimal generalized costs. Consequently, the methodology allows deriving the optimal price for a user-specified control horizon. Note that Table~\ref{tab:notation_method} denotes the used nomenclature for the presented methodology. 

\begin{table}[!t]
\centering
\begin{tabular}{lcl}
\toprule
Notation & Unit & Description \\
\midrule
     $k$        &  [s]  & Discrete simulation time step \\
     $T_{\mathrm{c}}$     &  [s]  & Control horizon \\
     $Q_I$      &  [veh/s] & Aggregated demand for region $I$ \\
     $N^*_I$ & [veh] & Optimal aggregated vehicle accumulation (DSO) for region $I$ \\
     $M_{IH}$ & [veh/s] & Transfer flow (QDUE) from origin $I$ to $H$  \\
     $M^*_{IH}$ & [veh/s] & Optimal transfer flow (DSO) from origin $I$ over $H$ \\
     $\theta^*_{IHJ}$ & - & Optimal route choice variable from region $I$ to $J$ via $H \in \mathcal{N}_I$ \\
     $N_{I,\mathrm{crit}}$ & - & Critical vehicle accumulation of region $I$ \\
     $N_{I,\mathrm{jam}}$ & - & Jam vehicle accumulation of region $I$ \\
     $n^*_{\mathrm{c},I}, n_{\mathrm{c},I}$ & - & Fraction of $N_I$ and $N_{I,\mathrm{crit}}$ in region $I$ (DSO, QDUE) \\
     VOT & [CHF/h] & Value of Time parameter \\
     $P^*_{IH}$     &  [CHF] & Pricing matrix for traversing from region $I$ to $H$ \\
     $\mathcal{M}$ & - & Set of pre-trained MLN network models  \\
     $\alpha_{II}, \alpha_{IJ}$ & - & Fraction of vehicle accumulation as LRHO model parameters \\
     $l$ & - & Index of peice-wise affine (PWA) function for MFD  \\
     $L$ & - & Total number of PWA functions \\
     $G^l_I(\cdot) $ & [veh/s] & PWA MFD function \\
     $f_{II}, f_{IH},  f_{HI}, f_{IHJ}$ & [veh/s] & LRHO decision variables for internal and transfer flows \\
     $N_p$ & - & Prediction horizon \\
     $k_p$ & [s] & Discrete prediction time step of LRHO \\
     
     $\tau_I, \tau_H, \tau_{IH}$ & [s] & Average travel time in region $I$, $H$, and $I$ via $H$ \\
     $\Tau_{IH}$ & [s] & Travel time matrix \\
     $c_{IH}, c^*_{IH}$ & [CHF] & Generalized and predicted costs traversing region $I$ via $H$ \\
     $C_{IH}^*, C_{IH}$ & [CHF] & Generalized cost matrix (DSO, QDUE) \\
     $U_{H,IJ}$ & - & Utility function from $I$ to $J$ and alternatives $H$ \\
     $\epsilon_H$ & - & Error term of unobserved determinations \\
     $\mu$ & - & MNL scaling parameter \\
     $p_{IH}$ & [CHF] & Optimal price traversing region $I$ via $H$ \\
     $P^*_{IH}$ & [CHF] & Optimal price matrix \\
     $x_{m}$ & - & Neuron input \\
     $y_{o}$ & - & Neuron output \\
     $w_{o,m}$ & - & Weight for neuron input \\
     $I_o$ & - & Summation of weighted input to neuron \\
     $F(\cdot)$ & - & Activation function of neuron \\
     TS$_I$ & [veh $\cdot$ h] & Time spent in region $I$ \\
     TTS & [veh $\cdot$ h] & Total Time spent in the network \\
     TTD & [veh $\cdot$ km] & Total traveled distance in the network \\
     MAE & [CHF] & Mean absolute error \\
     $N$ & [veh] & Vehicles served \\
     \bottomrule
\end{tabular}
\caption{Notation for the equilibria and pricing methodology.}
\label{tab:notation_method}
\end{table}

In more detail, the first block (Figure~\ref{fig:block_diagram_method}, top) represents the discretized multi-region model to simulate the vehicle accumulation trajectories $N_I(k)$, $N_{II}(k)$, and $N_{IJ}(k),~\forall k$, for a given exogenous demand scenario $Q_{IJ}(k),~\forall k$. Note that $k$ is the discrete-time step, and for $k=1$, i.e., the beginning of the simulation, an initial route choice with equal probability for all $\theta_{IHJ}(k=1)$ is set.  This is reasonable, as at $k=1$, the network is empty, and consequently, the costs for all paths are equal. Additionally, an initial price matrix $P^*_{IH}(k=1)$ denotes a user-defined starting price for traversing from $I$ via $H$. As this work considers dynamic congestion pricing, $P^*_{IH}(k=1)$ can be set to zero and will react accordingly due to the system feedback throughout time evolves.

The outputs $N_I(k)$, $N_{II}(k)$, $N_{IJ}(k)$, and $\theta_{IHJ}(k)$ are  used to determine the DSO and the QDUE, respectively. The equilibria constitute two important system states for defining the potential improvement the pricing methodology can add. As an output system states for $k+T_c$ are determined, which allows the computation of the splitting rates $\theta_{IHJ}(k+T_{\mathrm{c}})$, $\theta^{*}_{IHJ}(k+T_{\mathrm{c}})$, the transfer flows $M_{IH}(k+T_{\mathrm{c}})$, $M^{*}_{IH}(k+T_{\mathrm{c}})$ and the accumulations $N_{I}(k+T_{\mathrm{c}})$, $N^{*}_{I}(k+T_{\mathrm{c}})$ for the QDUE and DSO, respectively. Note that $T_{\mathrm{c}}$ denotes the control horizon and all optimal quantities, i.e., computed with DSO, are defined with an asterisk. Further, the block for equilibria derivation uses VOT$(k)$ as an input. The VOT is utilized here to calculate generalized costs of a trip through the multi-region network. Consequently, it serves as a parameter for the determination of DSO and QDUE splitting rates. The complete derivation of (a) the linear optimization problem formulation for DSO and the computation of QDUE with Dijstrka and MNL is introduced in Section~\ref{sec:eq_derivation}. 

Finally, the block for the pricing computation utilizes the route choice signals $\theta^*_{IHJ}(k+T_{\mathrm{c}})$ and $\theta_{IHJ}(k+T_{\mathrm{c}})$, and transfer flows $M^{*}_{IH}(k+T_{\mathrm{c}})$ and $M_{IH}(k+T_{\mathrm{c}})$ from DSO and QDUE, respectively. Also, the fractions of accumulations and critical accumulation ($N_{I,\mathrm{crit}}$) of a region $I$ serve as an input and represent the region's congestion level; the variable is denoted as $n_{\mathrm{c},I}$ and $n^*_{\mathrm{c},I}$. Note that all quantities are utilized as inputs to a pre-trained set of MLN network models $\mathcal{M}$, which allow the derivation of prices for horizon $k+T_c$ and every combination of origins $I$ and neighboring regions $H$. The determined prices are collected in a pricing matrix $P^*_{IH}(k+T_{\mathrm{c}})$, which is then applied to the simulation plant for the specified control horizon. The magnitude of every pricing matrix element reflects the required economic incentive to influence a travel option from $I$ to $J$ over $H$ for time horizon $k + T_{\mathrm{c}}$ towards the system optimal choice. The way we specify our machine learning models to derive the optimal generalized costs and calculate the optimal prices are introduced in detail in Section~\ref{sec:pricing}.

\subsection{EQUILIBRIA DERIVATION}
\label{sec:eq_derivation}
How users in a transportation network decide on their travel route is assumed to follow well-defined principles. The Wardrop principles proposed in~\cite{ref:Wardrop} describe the two states as the user equilibrium and the system optimum. As equilibria are not constant throughout time, to capture the traffic dynamics and the resulting varying route choice, the states are commonly denoted as DUE and DSO. The DUE constitutes that if users in a network are departing at the same time experience a minimal and equal travel time, the system operates in a DUE state~\citep{ref:YILDIRIMOGLU2014_DUE, ref:RAN1996_DUE}. In other words, every user tries to maximize their own utility i.e.; one tries to get from origin to destination as fast as possible. Also, this means that the travel time or the resulting travel costs are minimized, and no user can find a better minimal solution by adjusting the route choice. The derivation and analysis of the DUE have been extensively studied in works such as, e.g.,~\cite{ref:HUANG20201_DUE} or~\cite{ref:GUO2020_MFD_DUE}. Nevertheless, it has been shown that operating in the DUE does not lead to an overall maximization of system performance. Contrary to the DUE, the DSO shows a substantial improvement in network performance but also introduces for a subset of users longer routes to reach their destination. Thus, we introduce the derivation of both equilibria as fundamental work for the pricing models. \\

\underline{\textit{Derivation of DSO}}

At first, the linear derivation of DSO is introduced. The problem aims for deriving the optimal splitting rates $\theta^*_{IHJ}(k+T_{\mathrm{c}})$ for a given $T_{\mathrm{c}}$that consequently allow the derivation of the optimal internal flows $M^*_{II}(k + T_c)$, optimal transfer flows to neighboring regions $M^*_{IH}(k + T_c)$, and finally the optimal accumulation trajectories $N^*_{II}(k+T_{\mathrm{c}})$, $N^*_{IJ}(k+T_{\mathrm{c}})$, and $N^*_{I}(k+T_{\mathrm{c}})$. The multi-region model from Section~\ref{sec:model} is formulated with several nonlinearities (e.g.\ formulation of MFD function $G(\cdot)$, fraction of accumulations $N_{IJ}(t)/N_I(t)$, etc.). Hence, an NMPC is applied in several other studies focusing on optimal control~\citep{ref:isik_model_rg, ref:Tajalli, ref:Hajiahmadi}. This work formulates the problem as a linear model to allow the application of an LRHO; implying the utilization of a linear model. Therefore, the nonlinearities are removed by applying several approximations based on~\cite{ref:Genser2020TRB} and~\cite{kouvelas2017enhancing, ref:Kouvelas_Lin}.

First, the model parameters $\alpha_{II}(k)$ and $\alpha_{IJ}(k)$ are introduced, which are updated every time a predicted solution is applied to the simulation plant; i.e., the parameters remain constant over the prediction horizon and are updated when rolling the prediction horizon. $\alpha_{II}(k)$ and $\alpha_{IJ}(k)$ are defined as follows:
\begin{equation}
    \alpha_{II}(k) = \frac{N_{II}(k)}{N_I(k)}, ~ \forall I \in \mathcal{R}
\end{equation}
and 
\begin{equation}
    \alpha_{IJ}(k) = \frac{N_{IJ}(k)}{N_I(k)}. ~ \forall I,J \in \mathcal{R}
\end{equation}

Secondly, MFD functions $G_I(\cdot)$ are approximated with a number of piece-wise affine (PWA) functions; $l = \{1,2, ..., L\}$ denotes the index of PWA function and $L$ the total number of functions, chosen for an accurate approximation. In the following, each piece-wise linear MFD function is indicated by $G^l_I(\cdot)$. Thirdly, we introduce new decision variables:
\begin{equation}
\label{eq:fii}
    f_{II}(k) = \theta_{III}(k)G_I^l(N_I(k))\alpha_{II}(k), ~ \forall I \in \mathcal{R}
\end{equation}
and 
\begin{equation}
\label{eq:fih}
    f_{IH}(k) =  G^l_I(N_I(k)) \sum_{J \in \mathcal{R}} \theta_{IHJ}(k)\alpha_{IJ}(k), ~ \forall I,J \in \mathcal{R}, H \in \mathcal{N}_I 
\end{equation}
where $f_{II}(k)$ and $f_{IH}(k)$ define decision variables for internal and transfer flows, respectively. The right sides of equations (\ref{eq:fii}) and (\ref{eq:fih}) show the remaining nonlinearities by the product of $\theta_{III}(k)$ and $\theta_{IHJ}(k)$, respectively. The introduction of $f_{II}(k)$ and $f_{IH}(k)$ allow to complete the linearization of the problem. As in~\cite{ref:Kouvelas_Lin} the methodology was applied to find the optimal perimeter control, a transformation from $f_{II}(k)$ and $f_{IH}(k)$ to the original control variables is used. 

Nevertheless, variables $f_{II}(k)$ and $f_{IH}(k)$ only consider internal flows and transfer flows to a neighboring region $H$; i.e., the information about the final destination $J$ is not available. In our approach to determine the optimal splitting rates $\theta^*_{III}(k)$ and $\theta^*_{IHJ}(k)$ this information is necessary to ensure that the summation of flow proportions on every possible path from $I$ to $J$ is correct, as well as for the transformation to the original decision variables. Therefore, we introduce one additional decision variable $f_{IHJ}(k)$ that is constrained by
\begin{equation}
    \sum_{J \in \mathcal{R}} f_{IHJ}(k) = f_{IH}(k), ~ \forall I,J \in \mathcal{R}, H \in \mathcal{N}_I 
    \label{eq:constraint_fih}
\end{equation}
to ensure that splitting rates can be constrained correctly and calculation of the original decision signals $\theta^*_{III}(k)$ and $\theta^*_{IHJ}(k)$ can be obtained. Note that for $\theta^*_{III}(k)$ the result does not influence optimal route choices, as the splitting rate corresponds to users traveling from origin $I$, over $I$, to final destination $I$. Hence, all $\theta_{III}$ must be $\theta_{III}(k) = 1~\forall k$. Nevertheless, the decision signals are included in the algorithm and the results serve for validation purposes. Finally, we introduce an operational constraint to prevent the optimization results, i.e., the route choice signals, from oscillating:  

\begin{equation}
    \Big | \theta^*_{IHJ}(k) - \theta^*_{IHJ}(k-1) \Big | \leq \sigma ~ \forall I,J \in \mathcal{R}, H \in \mathcal{N}_I,
    \label{eq:constraint_op}
\end{equation}
where the left-hand side of the equation represents the absolute difference between the route choice signals in time and $\sigma$ denotes a user-defined parameter to constraint the magnitude deviation. 

An LRHO procedure is introduced and utilized to solve for optimal splitting rates $\theta^*_{IHJ}(k)$ for a prediction horizon of $N_p$:
\begin{align}
    \max_{N_{I}(k), f_{II}(k), f_{IH}(k)}& ~ T_c \cdot \sum^{k_p+N_p - 1}_{k=k_p} \sum_{I \in \mathcal{R}} \Big [ f_{II}(k) + f_{IH}(k) \Big ]
	\label{eq:lmpc_first} \\
	\mathrm{s.t.}&~ N_{I}(k+1) = N_{I}(k) + T_c \Big (Q_I(k) - f_{II}(k) - \nonumber \\\ &~\sum_{H \in \mathcal{N}_I} f_{IH}(k) + 
	\sum_{H \in \mathcal{N}_I} f_{HI}(k) \Big)\\\
	&~ \mathrm{equations~(\ref{eq:constraint_fih}),~(\ref{eq:constraint_op})} \\\
	&~ 0 \leq f_{II}(k) \leq \alpha_{II}G^l_{I}(N_I(k)) \\\
	&~ 0 \leq f_{IHJ}(k)  \\\
	&~ \sum_{H \in \mathcal{N}_I} f_{IHJ}(k) \leq \alpha_{IJ}(k) G^l_I(N_I(k)) \\\
	&~ 0 \leq N_I(k) \leq N_{I,\text{jam}} \\\
	&~ k = k_p, k_p + 1, ... , k_p + N_p -1 \label{eq:lmpc_last}\\\
	&~ \forall I,J \in \mathcal{R}, H \in \mathcal{N}_I \nonumber 
\end{align}
For every solution computed with the LRHO, a calculation of the optimal splitting rates $\theta^*_{IHJ}(k)$ can be performed by utilizing the variables $f_{IHJ}(k)$, $\alpha_{IJ}(k)$, and $G^l_{I}(N_I(k))$. Analogically, splitting rates for internal flows (as stated above the result has to be $\theta_{III}(k) = 1)$ can be evaluated with $f_{III}(k)$, $\alpha_{II}(k)$, and $G^l_{I}(N_I(k))$. Note that all constraints in equations (\ref{eq:lmpc_first})--(\ref{eq:lmpc_last}) are linear, and consequently, the problem can be solved with low computational power as a linear program. The results are utilized in Section~\ref{sec:pricing} as in input to the MLN network allowing for the derivation of the optimal generalized costs and the optimal pricing functions. \\

\underline{\textit{Derivation of QDUE}}\\
Whereas the network's throughput is maximized to compute the DSO, the DUE is defined by each user in the network, minimizing her/his travel costs (i.e., travel time). In this work, the DUE is approximated by finding the shortest paths from origin $I$ to a destination region $J$ with Djikstra algorithm and applying an MNL model. Consequently, we denote the equilibrium as QDUE. Modeling the inputs for these algorithms requires the costs of a trip in the network. Therefore, we first calculate the travel time of a trip from a region $I$ to a neighbor $H$ by utilizing the signals from the macroscopic model. The travel time $\tau_{IH}(k)$ can be defined as follows:
\begin{equation}
    \tau_{IH}(k) = \tau_I(k) + \tau_H(k) = \frac{\bar{L}_I \cdot N_I(k)}{G_I(N_I(k)) \cdot \bar{L}_I} + \frac{\bar{L}_H \cdot N_H(k)}{G_H(N_H(k)) \cdot \bar{L}_H}, \forall I\in \mathcal{R}, H \in \mathcal{N}_I 
\end{equation}
where $\tau_{I}(k)$ and  $\tau_{H}(k)$ are approximated by the fraction of average trip lengths $\bar{L}_I$, $\bar{L}_H$ and the corresponding estimated speeds (e.g., for $\tau_{I}(k)$  by utilizing the outflow $G_I(N_I(k))$, average trip length $\bar{L}_I$, and vehicle accumulation of a region $N_I$). Note that all elements for $\tau_{IH}(k)$ of a given network with arbitrary topology can be compiled in a travel time matrix $\Tau_{IH}(k)$.

To transform the elements of $\Tau_{IH}(k)$ into generalized costs that users experience when traveling through the network, we utilize the Value of Time (VOT). Hence, the generalized cost matrix $C_{IH}(k)$ can be defined by simply multiplying
\begin{equation}
    C_{IH}(k) = \Tau_{IH}(k) \mathrm{VOT}(k), ~ \forall I \in \mathcal{R}, H \in \mathcal{N}_I 
 \end{equation}
where VOT$(k)$ is the VOT for trips from $I$ to $H$ in the network. Generalized costs $C_{IH}(k)$ are then utilized to calculate the shortest paths with Djikstra algorithm (represent minimum users costs or maximum user utility) which are used as input alternatives for the MNL. The simulation model allows three path possibilities from $I$ to $J$ and therefore, three alternatives over neighboring region $H$ are allowed; $U_{H,IJ}(k)$ defines the utility function for individuals going from $I$ to $J$ via an alternative $H$. Essentially, $U_{H,IJ}(k)$ is modeled with a deterministic term, which is the corresponding element for a pair $(I,H)$ from the generalized cost matrix $C_{IH}(k)$; i.e., $U_{H,IJ}(k) = c_{IH}(k) + \epsilon_H$, where $c_{IH}(k) \in C_{IH}(k)$ and $\epsilon_H$ denotes an error term containing all unobserved determinations of the utility function; $\mu$ denotes a scaling parameter. Finally, the MNL is defined as follows: 
\begin{equation}
    \theta_{IHJ} (k) = \frac{\exp(\mu U_{H, IJ}(k))}{\sum_{H \in \mathcal{N}} \exp(\mu U_{H,IJ}(k))}. ~ \forall I,J \in \mathcal{R}, H \in \mathcal{N}_I 
    \label{eq:MNL}
\end{equation}  
The MNL definition is motivated by~\cite{ref:BenAkiva1999Logit} and allows the derivation of the QDUE splitting rates, which represent an approximation of the DUE. Finally, the sets of equilibrium quantities are derived and serve as input to the pricing models introduced in Section~\ref{sec:pricing}.

\subsection{Optimal pricing with Multi-Layer-Perceptron networks}
\label{sec:pricing}
This work utilizes the macroscopic multi-region model to derive optimal pricing for every region boundary in the network. Contrary to other works (e.g., ~\cite{ref:Gu2018JDTT} focusing on a Proportional-Integral (PI) scheme utilizing MFD to allow for maintaining a protected region at the critical vehicle accumulation; corresponding to maximum vehicular flow), here, we utilize the concept of machine learning and design MLN networks to calculate the unknown optimal generalized cost matrix. Thus, this methodology allows the optimal price computation for every region boundary and time step, reflecting the additional costs that a user should experience to lead the network to the optimal state (i.e., the DSO).

In Section~\ref{sec:methodology} we have derived the splitting rates of the QDUE and the DSO, respectively. As shown in equation~(\ref{eq:MNL}), the generalized cost matrix $C_{IH}(k)$ is utilized to compute the splitting rates $\theta_{IHJ}(k)$. However, for finding $\theta^*_{IHJ}(k)$, the LRHO procedure is applied, and thus, the optimal generalized costs $C^*_{IH}(k)$ are unknown. Also, the relationship from equation~(\ref{eq:MNL}) shows that recovering a generalized cost matrix (the quantity is part of the utility function) by knowing $\theta^*_{IHJ}(k)$ does not give a unique solution. Thus, a methodology is needed that models the reverse relationship between route choice and the generalized costs. 

For decades neural networks are applied in the transportation domain for different purposes, e.g., traffic flow prediction, traffic signal control, or license plate recognition. As a systematic review by ~\cite{ref:Wang_LSTM} shows, the different types of neural networks can be assigned to several different applications in transportation. For example, Convolutional Neural Networks (CNN) are widely used for visual recognition, such as vehicle detection or license plate recognition. Otherwise, Deep Neural Networks (DNN) or Recurrent Neural Networks (RNN) are utilized to master time series prediction or classification tasks. For more details the interested reader is referred to e.g.,~\cite{ref:Wang_LSTM} or~\cite{ref:Nguyen_deeplearning}. 

\begin{figure}[!b]
    \centering
    \includegraphics[width=0.5\textwidth]{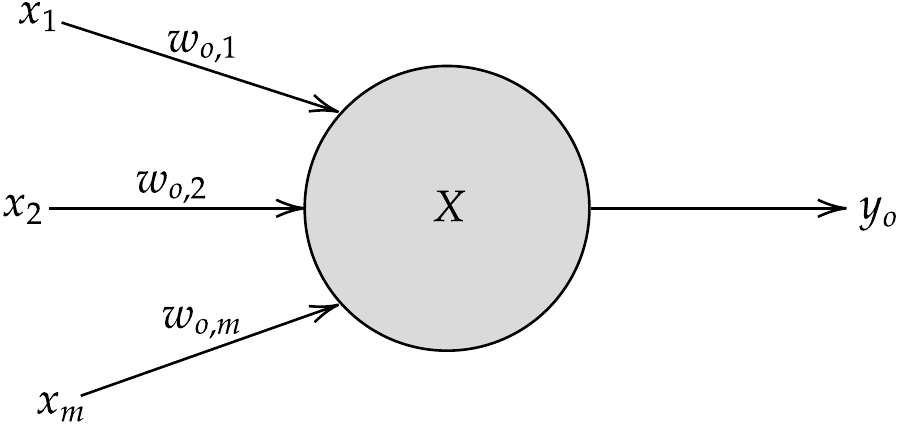}
    \caption{Schematic example of a neuron (based on~\cite{ref:Dougherty_NN}).}
    \label{fig:neuron}
\end{figure}

In this work, we design a specific type of DNN, a feed-forward MLN network. A MLN consists of one input layer, one hidden layer, and one output layer in its simplest form. The layers are composed of neurons taking one/multiple inputs and computing an output. The connections between neurons are modified by weights that scale the computations throughout the network. Figure~\ref{fig:neuron} depicts a neuron $X$; the illustration and mathematical description is based on~\cite{ref:Dougherty_NN}. The inputs are denoted by $\{x1, x2, ..., x_m\}$, the output by $y_o$, the weights for every input by $\{w_{o,1}, w_{o,2}, ..., w_{o,m}\}$. 

Mathematically, the output is computed by applying the following equations: 

\begin{equation}
    I_o = \sum_{i = 0}^{m} x_i w_{o,i}, 
\end{equation}
and 
\begin{equation}
    y_o = F(I_o),
\end{equation}
where $I_o$ denotes the summation of all weighted inputs and $F(\cdot)$ a transfer function (e.g., sigmoid, ReLU, etc.). 
By connecting a substantial number of neurons to a MLN network, such models allow the learning of complex non-linear relationships with the concept of supervised learning. For details, the reader is refereed to~\cite{ref:Dougherty_NN}.\\

\begin{figure}[t]
    \centering
    \includegraphics[width=0.8\textwidth]{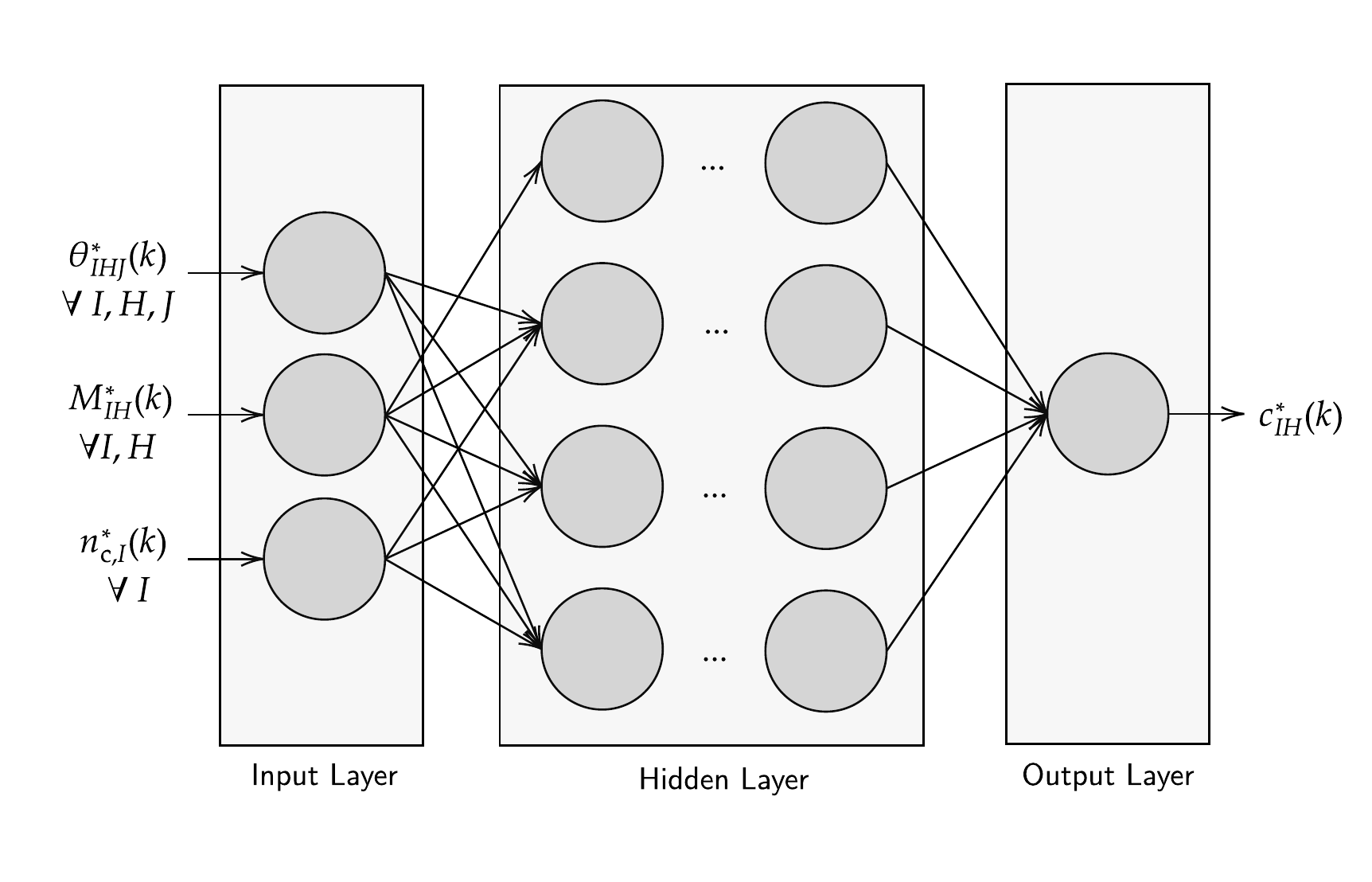}
    \caption{Neural network design for generalized cost estimation $c^*_{IH} \forall I,H$.}
    \label{fig:NN_sketch}
\end{figure}

\underline{\textit{MNL architecture}}\\
As depicted in Figure~\ref{fig:NN_sketch}, we train a set of MLN networks $\mathcal{M}$ to derive a unique pricing model for every region border in the network. Thus, every model takes as input features all elements of the splitting rate vector $\theta^*_{IHJ}~\forall I,H,J$, transfer flow vector $M^*_{IH}~\forall I,H$, and the vector containing $n^*_{\mathrm{c},I}~\forall I$. The output of every model is $c^*_{IH}$, which is an element of the optimal generalized cost matrix $C^*_{IH}$, i.e., $c^*_{IH} \in C^*_{IH}$. Also, the architecture of the models is sketched. The hidden layer is designed with two fully connected networks consisting of 50 hidden layers each. In both networks, the activation function ReLU is utilized. As an optimizer, the well-known Adam algorithm is applied, and as a loss function, the Mean Absolute Error (MAE) (defined in Section~\ref{sec:performance_metrics}) is used.\\

\underline{\textit{Datasets}}\\
 The training of pricing models requires (a) a QDUE data set where the generalized costs are known to tackle a regression problem with the backpropagation procedure and (b) the split of the data set into training data (70\% of the data set) and test data (30\% of the data). Note that the split of validation data is performed automatically by the utilized machine learning library. 
We utilize a dataset representing the QDUE simulation scenario, where all the required model inputs are available ($\theta_{IHJ}~\forall I,H,J$, $M_{IH}~\forall I,H$, $n_{\mathrm{c},I}~\forall I$).  
To prevent the models from learning demand-specific patterns (e.g., low demand or low vehicle accumulation at the beginning/end of the simulation), the data is randomly shuffled (a) before the split into training and test data and (b) during the training process. Also, the pre-processing ensures the scaling by applying a MinMaxScaler procedure, which is important as the splitting rates have a different variable range than the generalized costs.  \\

\underline{\textit{Hyperparameter tuning}}\\
Hyperparameter tuning is a crucial step when designing a neural network. Therefore, we tune the batch size and the number of epochs for model training. Besides, we investigate the learning rate of the stochastic gradient descent optimization algorithm. For a decrease of training time and an increase in model performance, an exponential decay function with an initial learning rate, a decay step, and a decay rate is utilized.  \\
 
\underline{\textit{Price function derivation}}\\
After predicting every element of $C^*_{IH}(k+T_{\mathrm{c}})$ for the control horizon $T_{\mathrm{c}}$, the final pricing matrix for every option from a region $I$ to $H$ (i.e., every region border) can be calculated by applying the following relationship: 

\begin{equation}
    P^*_{IH}(k+T_{\mathrm{c}}) = C_{IH}(k+T_{\mathrm{c}}) - C^*_{IH}(k+T_{\mathrm{c}}). ~ \forall I \in \mathcal{R}, H \in \mathcal{N}_I 
\end{equation}

Thus, every element of the pricing matrix $P^*_{IH}(k+T_{\mathrm{c}})$ contains the price $p^*_{IH}(k+T_{\mathrm{c}})$ for all implemented tolls in the multi-region-network and is applied for the future time steps of the simulation (Figure~\ref{fig:block_diagram_method}). 

\subsection{Performance metrics}
\label{sec:performance_metrics}
To evaluate the performance gain when pushing a system from the QDUE to DSO and to calculate measures for the effectiveness of our pricing methodology, we utilize a set of performance metrics. First, we define the Time Spent (TS) in [veh$\cdot$h] for a region $I$ as follows:  
\begin{equation}
    \mathrm{TS}_I = \sum_{k = 1}^{T} N_I(k). 
\end{equation}
Further we define the Total TS (TTS) to also evaluate the performance of the whole multi-region network. TTS in [veh$\cdot$h] is defined by

\begin{equation}
    \mathrm{TTS} = \sum_{k = 1}^{T} \sum_{I \in \mathcal{R}} N_I(k) = \sum_{I \in \mathcal{R}} TS_I.
\end{equation}
Another aggregated performance metric often used for macroscopic models is the Total Traveled Distance (TTD). The metric is computed by utilizing the average trip length $\bar{L}_I$, the internal flows $M_{II}(k)$ of a region $I$ and the transfer flows $M_{IHJ}(k)$ that leave region $I$ via a neighbor region $H$ to a destination region $J$. Note that $J \neq I$ must hold. The following equation defines TTD: 

\begin{equation}
    \mathrm{TTD} = \sum_{k = 1}^{T} \sum_{I \in \mathcal{R}} \bar{L}_I \Big (M_{II}(k) + \sum_{H \in \mathcal{N}_I} \sum_{J \in \mathcal{N_I}; J \neq I} M_{IHJ}(k)\Big).
\end{equation}

The consistency of all the simulations, the optimization procedure, and the pricing methodology is evaluated by the number of vehicles served, which must be consistent throughout all simulation scenarios. The number of vehicles served is computed with the following equation: 

\begin{equation}
    N = \sum_{k = 1}^{T} \sum_{I \in \mathcal{R}} N_I(k).
\end{equation}
Finally, to verify the learning process of the MLN networks and detect the overfitting of models, we utilize the MAE (also denoted as loss) defined as follows: 

\begin{equation}
    {\rm MAE} = \frac{1}{T}\sum_{k=1}^{T} \Big |c^*_{IH} - c_{IH} \Big |,
    \label{eq:MAE}
\end{equation}
where $c^*_{IH}$ represents the predicted generalized cost and $c_{IH}$ the generalized costs from the test dataset. $k$ is here utilized to sum the errors over $T$.
The next section will utilize all the performance metrics to evaluate the numerical experiments from a case study in Zurich, Switzerland.  

\section{NUMERICAL EXPERIMENT AND RESULTS}
\label{sec:casestudy}

\subsection{Simulation set-up and scenario design}
\label{sec:scenario-design}
\begin{figure}[!b]
    \centering
      \begin{subfigure}[b]{0.45\textwidth}
       \includegraphics[width=1\textwidth]{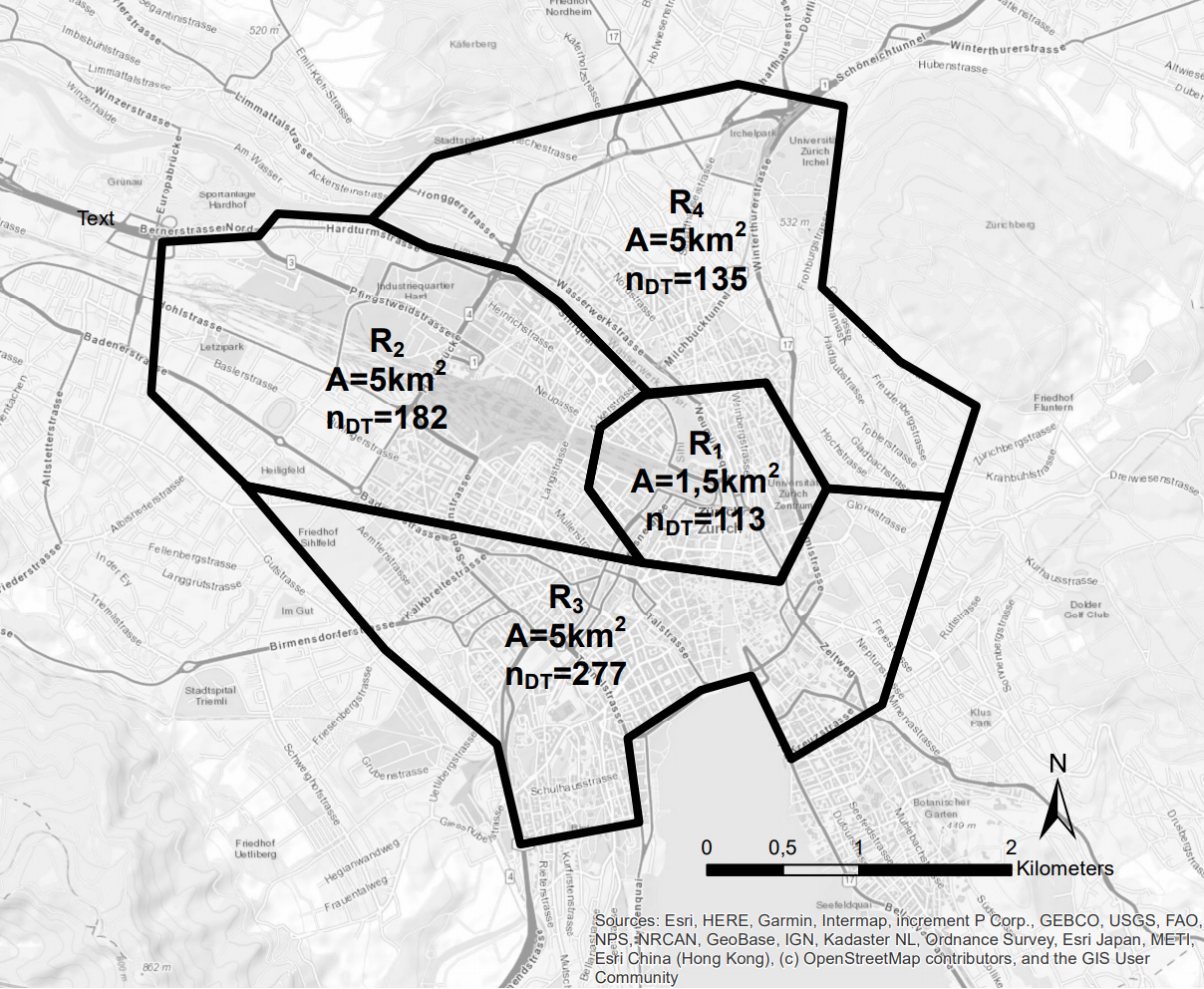}
        \caption{}
        \label{fig:regions}
    \end{subfigure}
    \begin{subfigure}[b]{0.45\textwidth}
        \centering
        \includegraphics[width=0.9\textwidth]{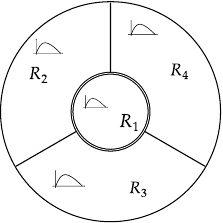}
        \caption{}
        \label{fig:region_model}
    \end{subfigure}
    \caption{Regions design for the city of Zurich and multi-region-network model; (a) every region is stated with an ID ($R_1$ -- $R_4$), area $A$, and number of available LDs $n_{\mathrm{DT}}$; (b) region $R_1$ is modeled as the city center (indicated by the double lines); $R_2$, $R_3$, and $R_4$ represent the boundaries to the city center.}\label{fig:region_mfd}
\end{figure}

This section presents a numerical experiment, where modeling is based on an example of the city of Zurich. The region design is derived from analyzing the main traffic arterials of Zurich and geographical reference of available Loop Detectors (LD). The city center (denoted as $R_1$) corresponds to an area of 1.5 [km$^2$] and has 113 LDs available. Consequently, parameters for MFD design are assumed with realistic values as follows: Jam accumulation $N_{1,\text{jam}}$ = 5000 [veh], average trip length $\bar{L}_1$ = 500 [m], and a network length $L_{1,n}$ of 30 lane kilometers. $R_2$, $R_3$, and $R_4$ denote the neighboring regions of the city center and are designed with an area of 5.0 [km$^2$] each. The number of detectors for regions $R_2$, $R_3$, and $R_4$ are 182, 277, and 135, respectively. MFDs for the border regions is designed with $N_{2,\text{jam}}=N_{3,\text{jam}}=N_{4,\text{jam}}$ = 8000 [veh], $\bar{L}_{2}=\bar{L}_{3}=\bar{L}_{4}$ = 2000 [m] and a road length $L_{2,n}=L_{3,n}=L_{4,n}$ of 48 lane kilometers, respectively. Hence, the entire network is designed for a storage capacity of 29000 vehicles. The region design is depicted in Figure~\ref{fig:regions} and Table~\ref{tab:parameters} shows the introduced region parameters. 

\begin{table}[!b]
	\centering
	\caption{Parameters for the design of the city center ($R_1$) and the border regions ($R_2$ -- $R_4$)}
	\label{tab:parameters}
	\begin{tabular}{lllrr}
	\toprule
    Parameter     & Variable           & Unit      & City center $R_1$ & Border regions $R_2$ -- $R_4$ \\
    \midrule
    Area & $A$                   & {[}km$^2${]} & 1.50           & 5.00           \\
    Number of LDs & $n_{\mathrm{DT}}$        & {[}-{]}   & 113            & 182, 277, 135    \\
    Jam accumulation  & $N_{I,\text{jam}}$   & {[}veh{]} & 5000           & 8000           \\
    Average trip length & $\bar{L}_I$ & {[}m{]}   & 500           & 2000           \\
    Network length & $L_{I,n}$    & {[}m{]}   & 30.000         & 48.000     \\ 
    \bottomrule
    \end{tabular}
\end{table}

Considering parameters design, we are proposing a four region network (Figure~\ref{fig:region_model}), where region $R_{1}$ represents the city center. The derived inputs to determine MFDs for $R_1$ -- $R_4$ are listed in Table~\ref{tab:input_mfd}. The maximum outflows $q_{\text{out}}$ are considered as 4.50 [veh/s] and 6.00 [veh/s] for the city center $R_1$ and the border regions $R_2$, $R_3$ and $R_4$, respectively. Jam density $\kappa$ is derived by $N_{I,\text{jam}}/L_{I,n}$ and assigned equal to 0.16 [veh/m] for all regions. 

\begin{table}[!t]
	\centering
	\caption{Parameters for the MFD design of the city center ($R_1$) and the border regions ($R_2$ -- $R_4$)}
	\label{tab:input_mfd}
	\begin{tabular}{lrrr}
	\toprule
	Parameter         & Unit                 & City center $R_1$ & Border regions $R_2$ -- $R_4$ \\
	\midrule
	$q_{\text{out}}$ 		  &[veh/s]      & 4.50             & 6.00                      \\
	$\kappa$  & [veh/m]     & 0.16             & 0.16                      	\\
	$a$	  &[-]        & 2.10 $\cdot 10^{-10}$             & 7.72 $\cdot 10^{-11}$               \\
	$b$ 		  & [-]       & -2.25 $\cdot 10^{-6}$             & -1.25 $\cdot 10^{-6}$                      \\
		$c$	  &[-]        & 6.06 $\cdot 10^{-3}$             & 5.13 $\cdot 10{^-3}$                 \\
	\bottomrule
	\end{tabular}
\end{table}

\begin{figure}[!b]
      \centering
       \includegraphics[width=0.6\textwidth]{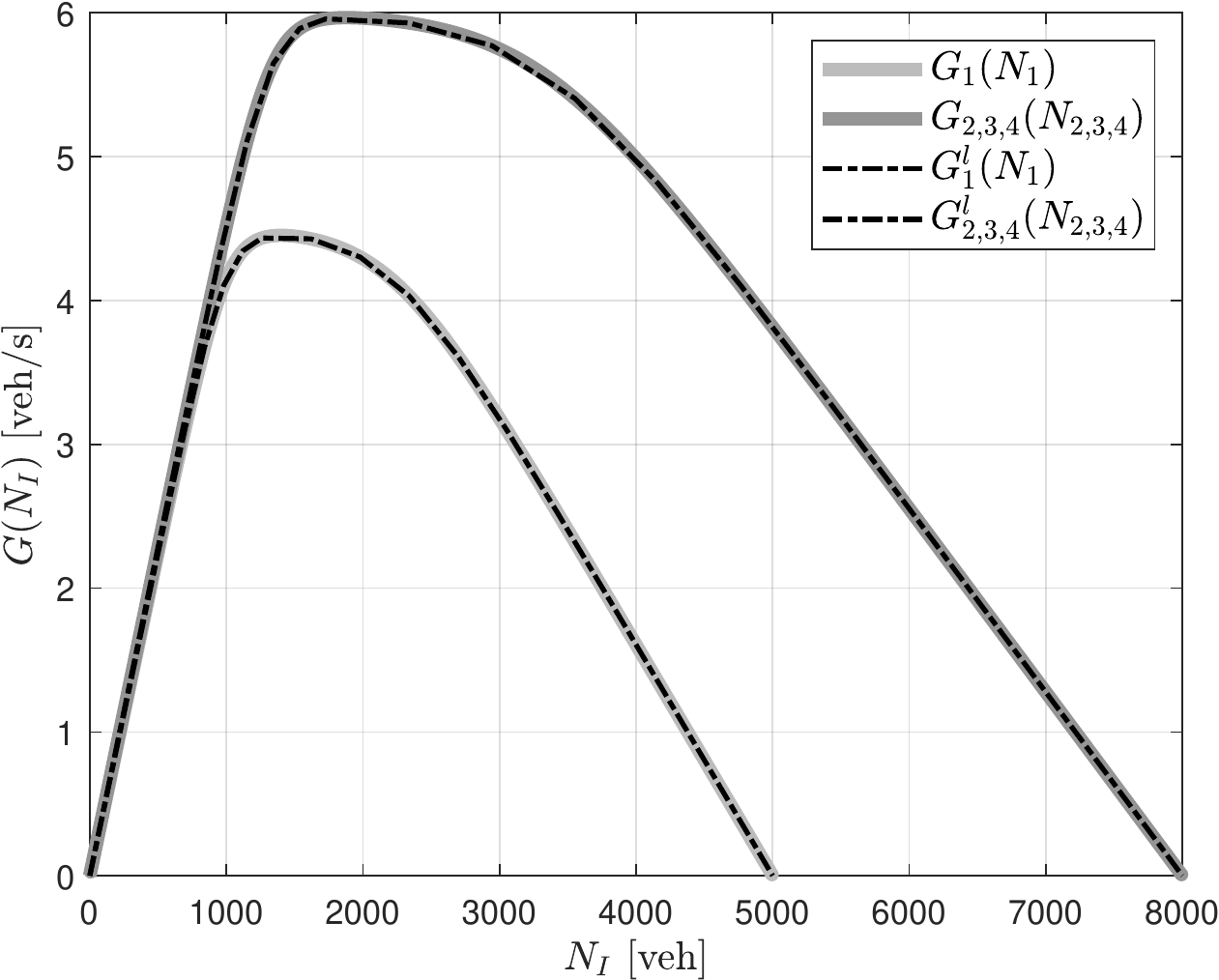}
    \caption{The MFDs are designed according to assumptions related to the City of Zurich (region size, partitioning, etc.) and one can note that $R_2$, $R_3$ and $R_4$ are modeled as larger regions with higher capacity. The dashed lines are representing the linear fit of each MFD, respectively.}
     \label{fig:mfds}
\end{figure}

Note that parameters $a$, $b$, and $c$ correspond to the parameters of the polynomial MFD representation and do not have a physical meaning. Figure~\ref{fig:mfds} depicts the designed MFDs with the parameters from Table~\ref{tab:input_mfd}. Also, the affine approximations for the city center and border regions are shown, respectively. These functions $G_1^l(N_1)$ and $G_{2,3,4}^l(N_{2,3,4})$ are utilized for the DSO calculation later. The approximation granularity for each MFD is specified by the number of lines $l=20$, considering the computational effort and minimization of the approximation error.

\begin{figure}[!t]
    \centering
    \includegraphics[width=0.6\textwidth]{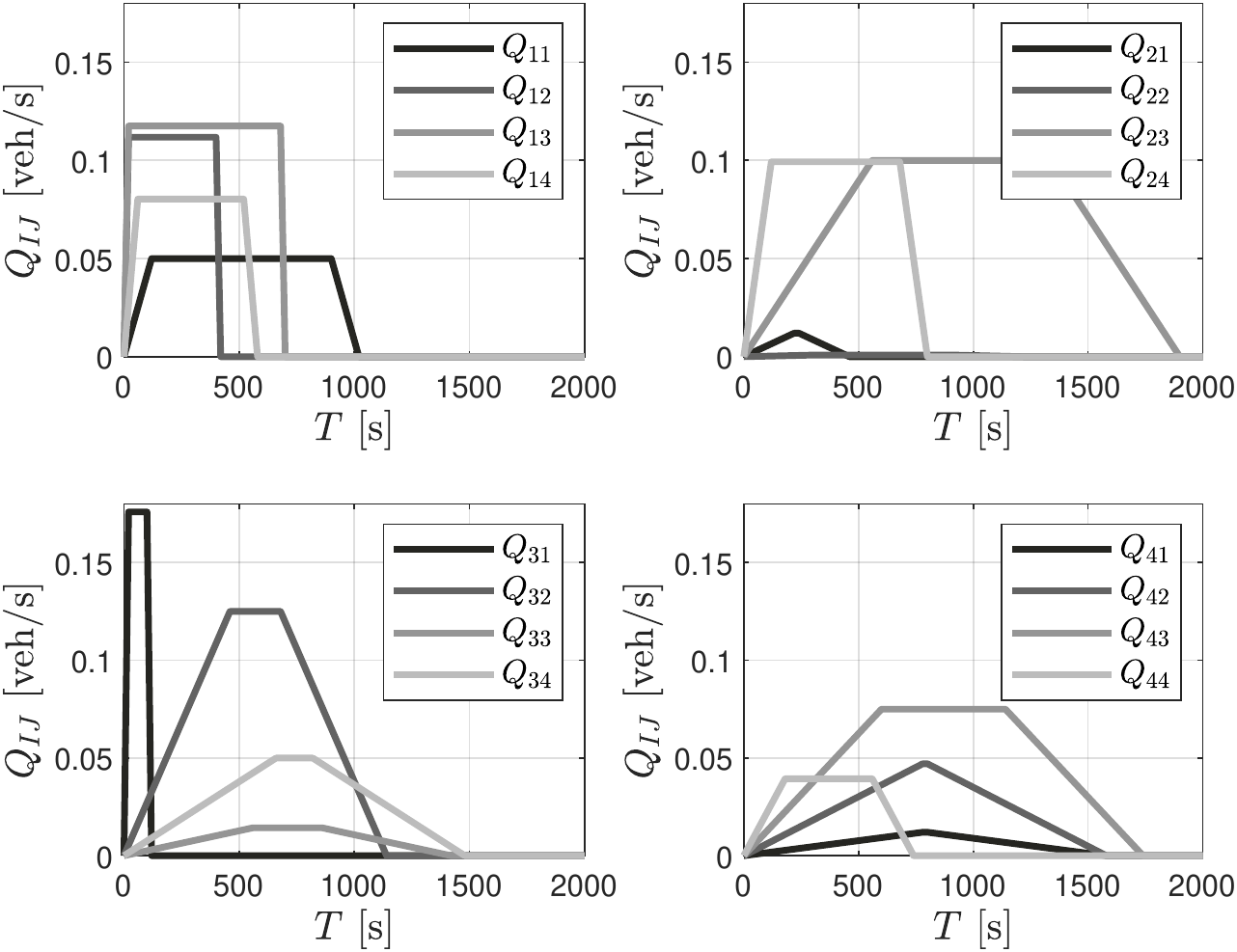}
    \caption{Traffic demand per region and pre-defined simulation horizon; configuration is for a 4X4 OD matrix, where $I$ specifies to the origin and $J$ the destination.}
    \label{fig:demand_patterns}
\end{figure}

As a relevant peak-hour simulation scenario for testing the pricing methodology is required, representative demand patterns are derived. Therefore we defined target accumulations for every region and determined representative trapezoid parameters by solving an optimization problem. $R_1$ represents a traffic situation in the congested regime, whereas $R_2$ and $R_4$ operate always in non-congested states. $R_3$ reaches the critical vehicle accumulation for a short time but shows no severe congestion. Furthermore, the demand magnitudes show that $R_1$ and $R_2$ are contributing more to the accumulation trajectories, compared to $R_3$ and $R_4$. The derived demand patterns $Q_{IJ}$ [veh/s] are depicted in Figure~\ref{fig:demand_patterns}. Note that after 2000 [s] of simulation time, no demand is present to ensure that we can clear the network at the simulation end; i.e., all users can finish their trip until 3000 [s] of simulation. 

\subsection{Equilibria results and comparison}
\label{sec:equilibria_results}
\begin{figure}[!t]
    \centering
    \includegraphics[width=0.6\textwidth]{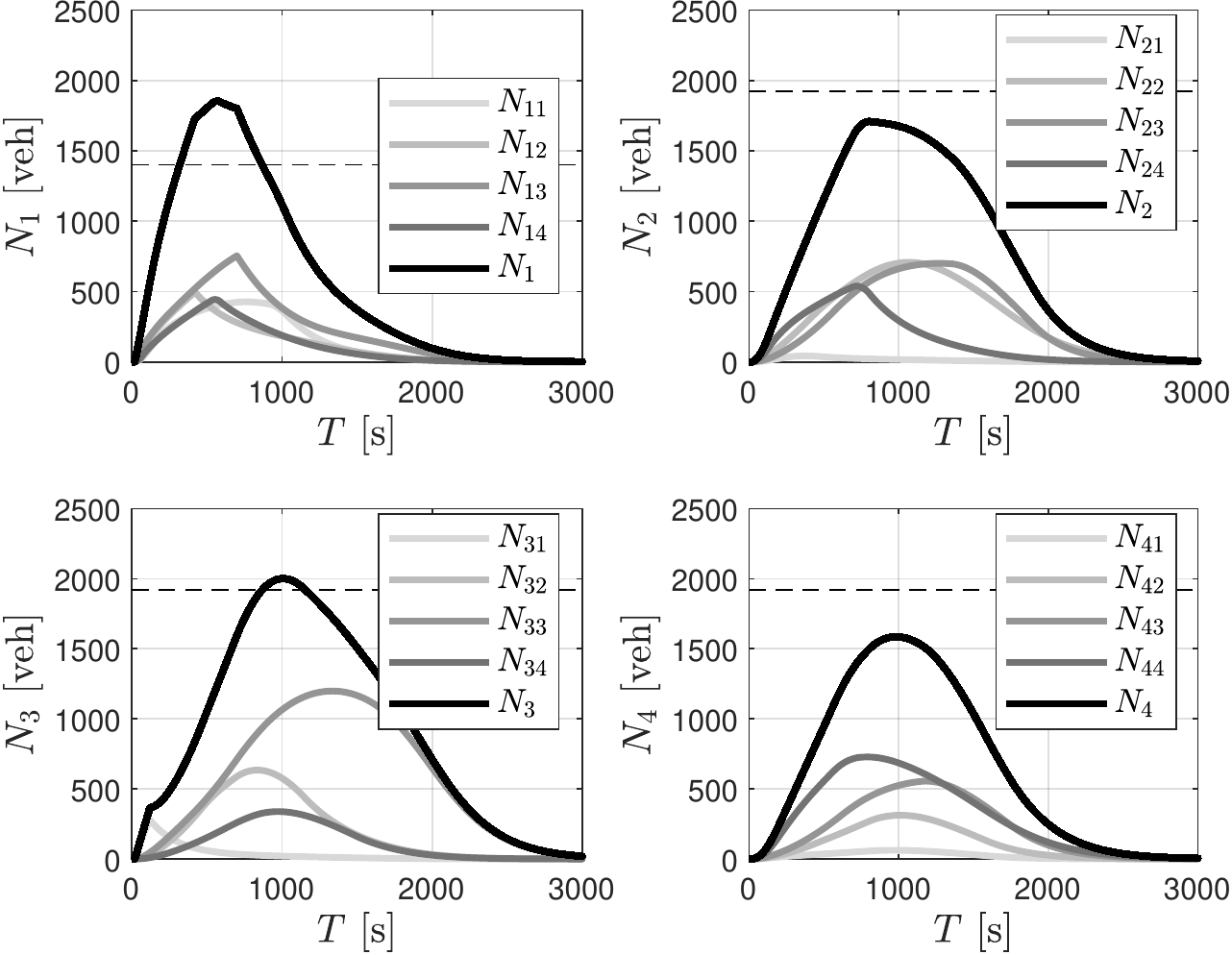}
    \caption{Accumulation trajectories $N_{IJ}$ for the QDUE scenario of $R_1$ -- $R_4$.}
    \label{fig:DUE_accumulation}
\end{figure}

The created demand scenario serves as an exogenous input to the simulation plant. Thus, the accumulation trajectories for every $N_{IJ}(k)~\forall k$ and all route choice signals $\theta_{IHJ}(k)~\forall k$ are computed for a simulation horizon of 3000 [s]; this time horizon was chosen to ensure an empty network for the given traffic demand. The transformation of travel times in the multi-region model into generalized costs is performed by applying a VOT of 27 [CHF/h] (based on the study from~\cite{ref:Hoerl2019VOT}). The simulated scenario represents the QDUE. Figure~\ref{fig:DUE_accumulation} depicts the vehicle accumulation trajectories for all $N_{IJ}$. The solid black trajectory in every subplot represents the aggregated accumulation $N_I$ for regions 1 -- 4. The critical vehicle accumulations $N_{I,\mathrm{crit}}$ for all regions are depicted with the horizontal dashed line. As expected by design, $R_1$ experiences substantial congestion whereas $R_2$ and $R_4$ operate in free-flow conditions respectively. $R_3$ reaches the critical vehicle accumulation of 1920 [veh] around 1000 [s] of simulation time. 

\begin{figure}[!t]
    \centering
    \includegraphics[width=0.6\textwidth]{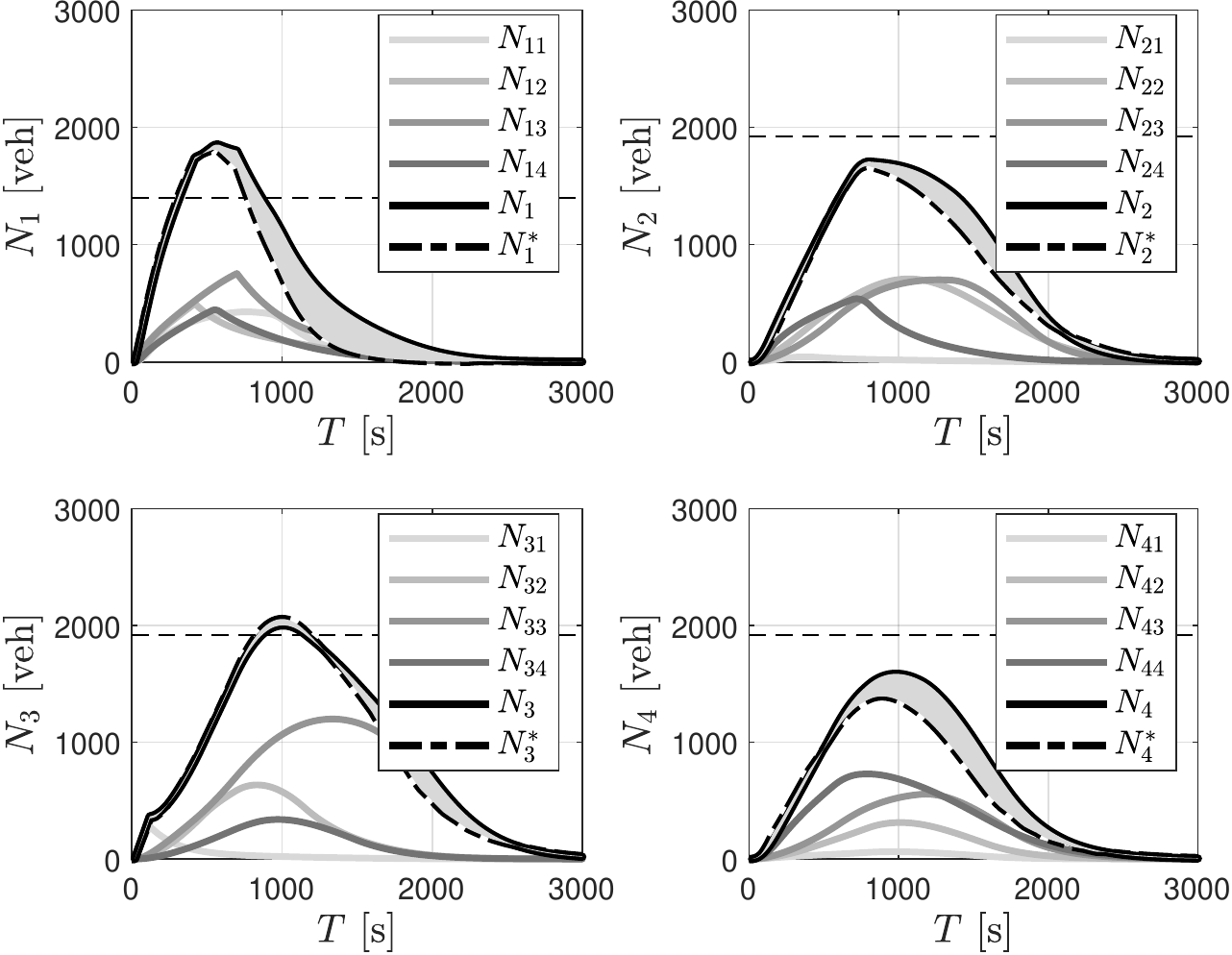}
    \caption{Accumulation trajectories $N^*_{IJ}$ for $R_1$ -- $R_4$ operating in the DSO. Note that the dashed line represents the aggregated vehicle accumulation $N_I$ of the DSO and the highlighted area the performance improvement.}
    \label{fig:acc_DSO_QDUE}
\end{figure}

To derive the performance gain of DSO, we use the linear model approximation and LRHO method to compute the optimal splitting rates  $\theta^{*}_{IHJ}(k)$. The linear program aims at maximizing the flow in the multi-region-network and was designed with the following parameters: prediction horizon $N_p = 3$; control cycle $N_c = 4$; control time step $T_c = 20$sec; operational parameter $\sigma = 0.2$. $N_p$ and $N_c$ are chosen concerning computational complexity and system response. Finding the optimal splitting rates $\theta^*_{IHJ}(k)$ allow to simulate the optimal accumulation trajectories $N^*_I(k)$, $N^*_{II}(k)$, and $N^*_{IJ}(k)$. Figure~\ref{fig:acc_DSO_QDUE} shows the vehicle accumulation for the optimal traffic distribution in the four-region network (dashed aggregated lines). Additionally, the aggregated accumulations of the QDUE for every region $I$, i.e., $N_I$ are shown again for comparison (solid black lines). The highlighted area between the aggregated trajectories denotes the performance improvement in vehicle accumulation between the QDUE and the DSO for all regions, respectively. 

\begin{figure}[tb]
    \centering
    \includegraphics[width=0.6\textwidth]{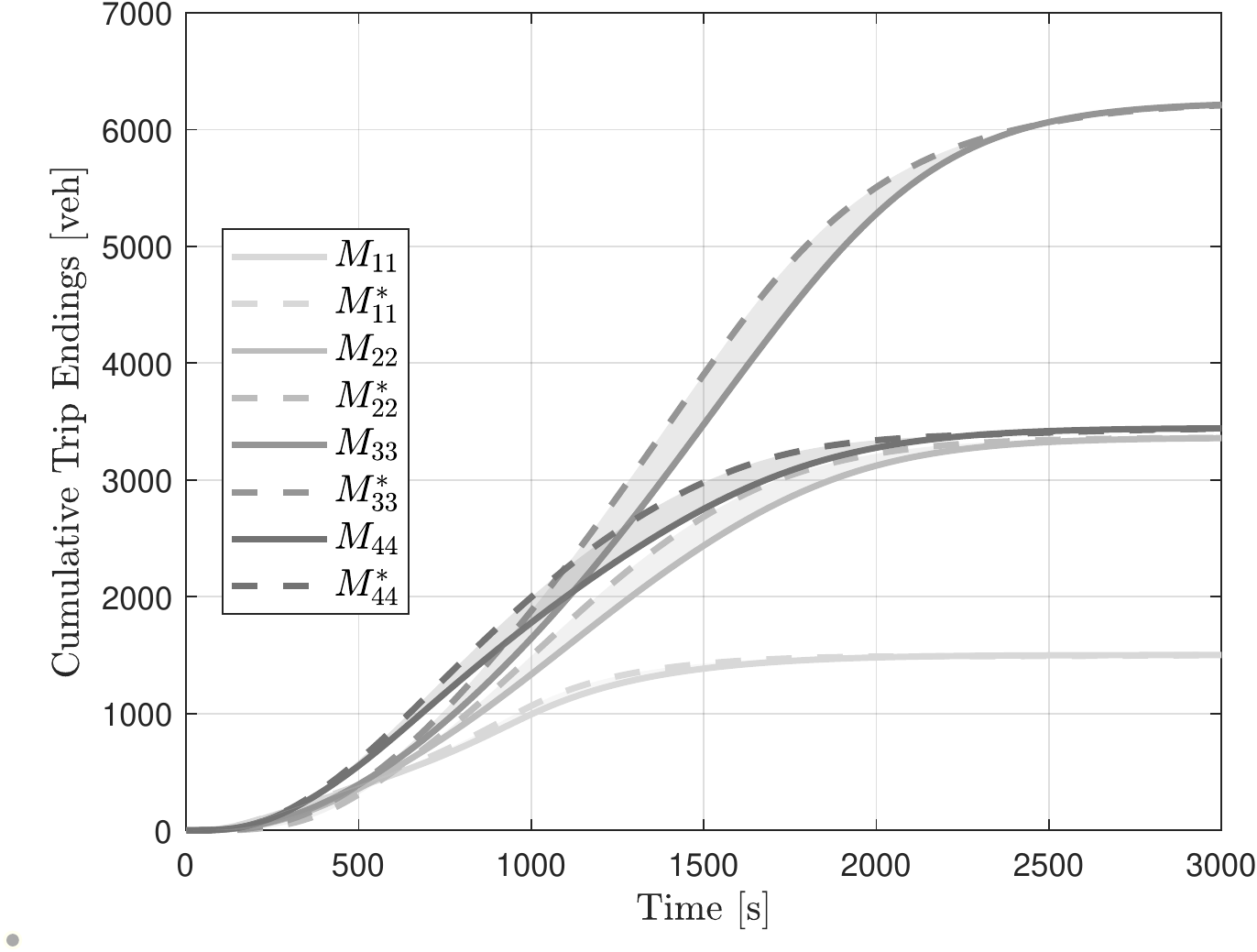}
    \caption{Cumulative trip endings $M_{II}$ and $M^*_{II}$ for the QDUE and DSO for $R_1$ -- $R_4$. The enclosed areas denote the performance improvement}
    \label{fig:cumulative_trip_endings}
\end{figure}

The accumulation trajectories of the DSO show that in every region, vehicle accumulation is reduced. In region $R_1$, where the network experiences congestion, an operation in DSO allows mitigation of congestion. Although $N^*_1$ exceeds the critical vehicle accumulation for a short time, the congestion dissolves faster than in the baseline scenario ($N_1$). The same behavior is shown in $R_2$. In $R_4$, the vehicle accumulation peak around 1000 [s] of simulation time can be reduced significantly. Only in $R_3$ the accumulation peak slightly higher than in the QDUE. However, an improvement is shown when the exogenous demand and consequently vehicle accumulation gets lower. An inspection of the cumulative trip endings for every region $R_1$-- $R_4$ additionally shows the performance improvement (Figure~\ref{fig:cumulative_trip_endings}). The area between every pair $(M_{II}, M^*_{II})$ indicates that in all regions, users can finish their trip earlier; i.e., the experienced reduction in delay is depicted. Also, both equilibria show the same magnitude of final cumulative trip endings, proving that the optimization problem is formulated correctly. 

A quantitative analysis of the performance metrics TS$_I$, TTS, TTD, and $N$ between the QDUE and DSO are compiled in Table~\ref{tab:performance_comparison}. When the system operates in the QDUE and splitting rates are determined by Dijkstra algorithm and MNL the following results are determined: The TS of $R_1$ is 17.97 [veh$\cdot$h$\cdot$10$^5$] and of $R_2$ 23.03 [veh$\cdot$h$\cdot$10$^5$]. The results for the DSO show a reduction to 14.03 [veh$\cdot$h$\cdot$10$^5$] and 21.01 [veh$\cdot$h$\cdot$10$^5$] for $R_1$ and $R_2$; corresponding to an improvement of 21.93\% and 8.77\%, respectively. The border regions $R_3$ and $R_4$ show a TS of 27.72 [veh$\cdot$h$\cdot$10$^5$] and 19.42 [veh$\cdot$h$\cdot$10$^5$], respectively. In the DSO, the metrics reduce to 26.50 [veh$\cdot$h$\cdot$10$^5$] and 16.51 [veh$\cdot$h$\cdot$10$^5$] which corresponds to an improvement of 4.39\% and 15.00\%, respectively. Note that $R_3$ shows the lowest performance improvement as depicted in Figure~\ref{fig:acc_DSO_QDUE}. Finally, the aggregated performance is determined with 88.15 [veh$\cdot$h$\cdot$10$^5$] for the QDUE and 78.06 [veh$\cdot$h$\cdot$10$^5$] for the DSO; resulting in an performance improvement of 11.45\% of TTS. 
We also compare the TTD for the whole network: For the QDUE the TTD is computed with 54.23 [veh$\cdot$km$\cdot$10$^6$]; DSO shows a TTD of 49.96 [veh$\cdot$km$\cdot$10$^6$] corresponding to an improvement of 7.87\%. 
Finally, we also compute the number of vehicles served $N$ for both equilibria. The computation supports the consistent trip endings in Figure~\ref{fig:cumulative_trip_endings}: 14.54 [veh$\cdot$10$^3$] are served in the QDUE and also in the DSO.

\begin{sidewaysfigure}[htbp]
    \includegraphics[width=1\textwidth]{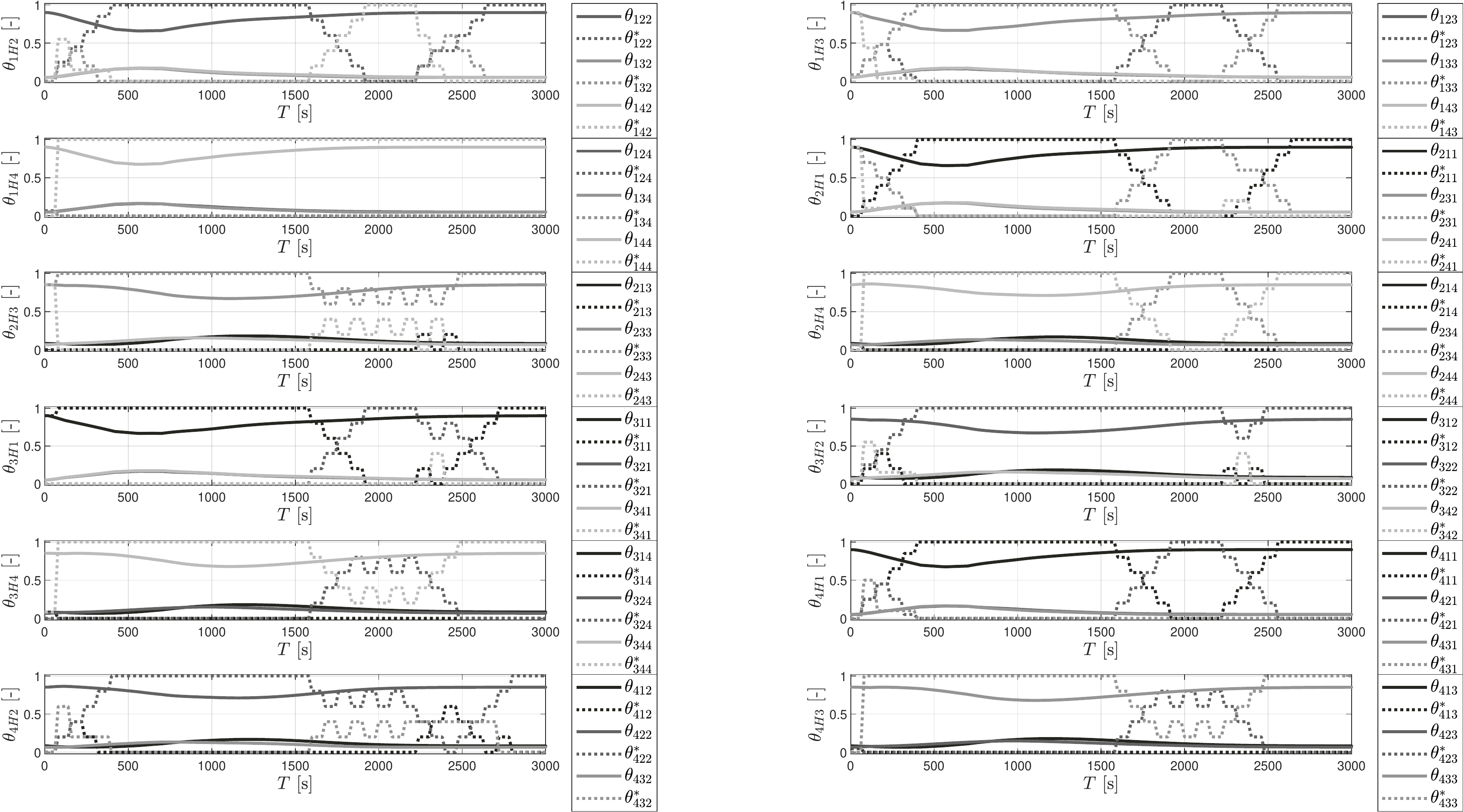}
    \caption{Route choice signals of the QDUE $\theta_{IHJ}$ (solid lines) and the DSO $\theta^*_{IHJ}$ (dashed lines), respectively.}
    \label{fig:thetas_comparison}
\end{sidewaysfigure}

\begin{table}[!t]
\centering
\caption{Comparison of performance metrices for the QDUE and DSO, respectively. Performance improvements (stated as Impr.) and the difference of vehicles (stated as Diff.) are denoted separately.}
\begin{tabular}{lrrr}
\toprule
  & QDUE   &  DSO &  Impr./Diff. \\
  & [veh$\cdot$h$\cdot$10$^5$] &  [veh$\cdot$h$\cdot$10$^5$] & [\%] \\
 \midrule
TS$_1$             & 17.97           & 14.03         & 21.93         \\
TS$_2$             &  23.03      &    21.01         & 8.77\\
TS$_3$            & 27.72   &   26.50   &  4.39        \\
TS$_4$            & 19.42   &  16.51    & 15.00 \\  
TTS           & 88.15   &  78.06   &   11.45 \\
\midrule
\midrule
 & QDUE   &  DSO &  Impr./Diff. \\
 & [veh$\cdot$km$\cdot$10$^6$] &  [veh$\cdot$km$\cdot$10$^6$] &  [\%] \\
 \midrule
 TTD          &  54.23  &  49.96    & 7.87  \\
\midrule
\midrule
 & QDUE   &  DSO &  Impr./Diff. \\
 & [veh$\cdot$10$^3$] & [veh$\cdot$10$^3$] & [veh] \\
 \midrule
$N$           & 14.54         & 14.54    & 0  \\
\bottomrule
\end{tabular}
\label{tab:performance_comparison}
\end{table}

Finally, we show the splitting rate signals for the QDUE and DSO in Figure~\ref{fig:thetas_comparison}, i.e., $\theta_{IHJ}$ (full lines) and $\theta^*_{IHJ}$ (dashed lines) for all allowed combinations of both variables. The difference in route choice between the two equilibria can be observed especially between 500 seconds and 1500 seconds of simulation time, where a certain high vehicle accumulation is present. In the QDUE all of the route choice signals show one path with the highest probability of being chosen. During high vehicle accumulations, this probability reduces, and the other two paths become more likely to be chosen. Contrary, in the DSO, traffic is guided via one path with the highest probability of $\theta^*_{IHJ} = 1$.
Additionally, it can be observed that after 1500 seconds of simulation time, the DSO route choice signals change more frequently than in the QDUE. This holds for all path possibilities except $\theta_{1H4}$. Note that the smoothness of the signals differs because of the control horizon $T_c$; i.e., the routing signals of the DSO only change every 80 seconds after a period of $N_c$; whereas the QDUE signals are updated for every simulation time step of 20 seconds. The difference in the signals indicate a significantly different routing of users through the network. Hence, it is shown that there must be a particular incentive for users to switch their route from QDUE to DSO. Finally, note that the DSO signal only changes after a period of the control horizon $N_c$ passed and the solution of the LRHO problem is applied.

\subsection{Training and application of pricing models}
\label{sec:training_app_pricing_models}
Both equilibria serve as an input for training the MNL models and for online prediction of the generalized costs. First, we take the QDUE scenario and extract the variables which are needed to create the dataset. Hence, we utilize the splitting rates $\theta_{IHJ}$, the transfer flows $M_{IH}$, and the fraction of the vehicle and critical accumulation $n_{\mathrm{c},I}$ of all regions. 
We compile a dataset with 150 data samples, i.e., a training data set with 105 samples (70\% of the data) and a test data set with 45 samples (30\% of the data).  

\begin{figure}[!t]
    \centering
    \includegraphics[width=0.6\textwidth]{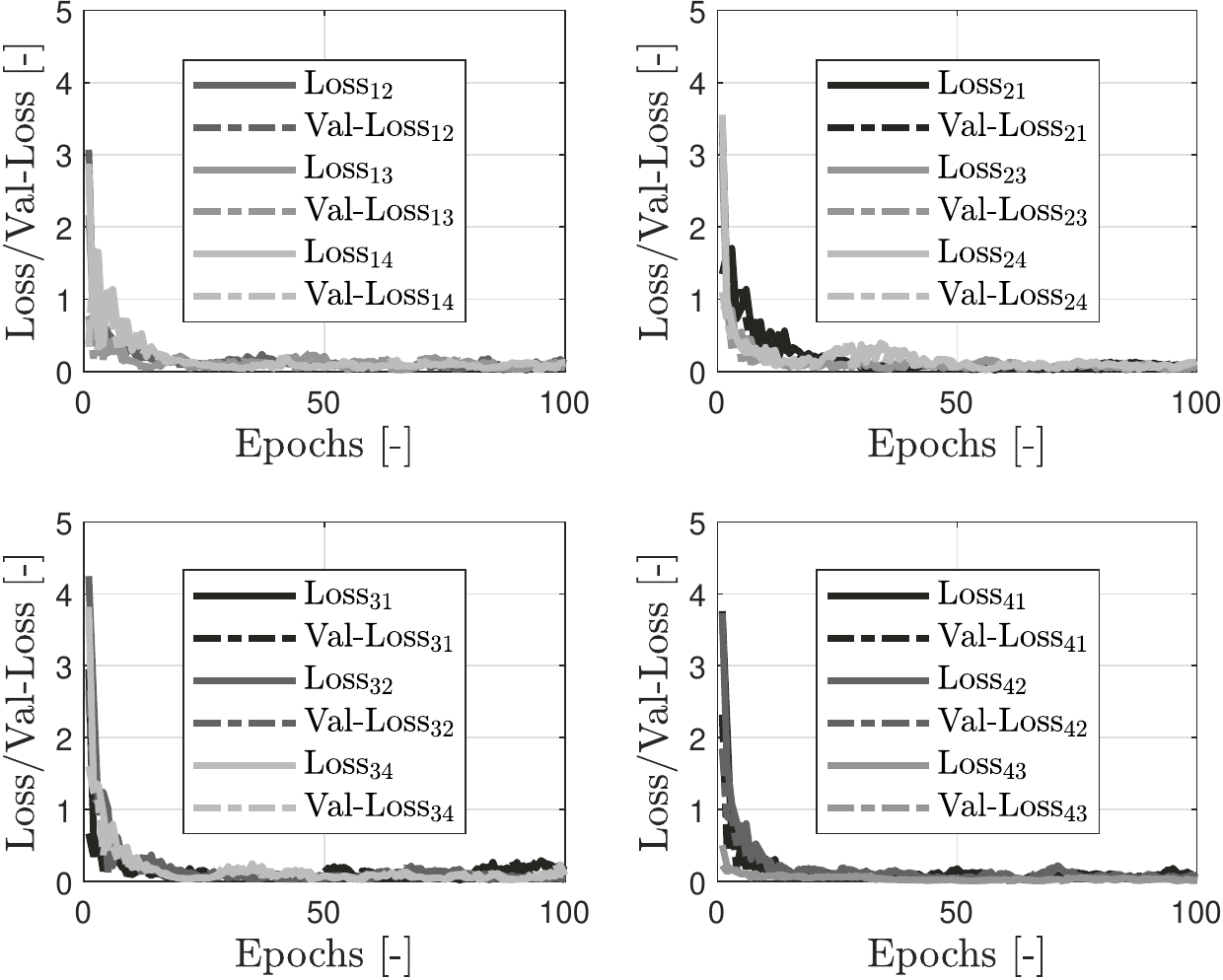}
    \caption{Loss and validation-loss for all tolls where pricing with $p^*_{IH}$ is performed.}
    \label{fig:loss}
\end{figure}

We utilize the designed MNL architecture of two fully connected networks with 50 layers each. The training of each model is performed with 100 epochs, a batch size of 64, and a validation split of 20\%. Additionally, the learning rate with an exponential decay function is utilized with the following parameters: initial learning rate of 0.01, decay-steps of 10000, and a decay rate of 0.9. The optimizer Adam is used with standard settings, and the loss function MAE is utilized. Note that the listed parameters result from the hyperparameter tuning procedure. Figure~\ref{fig:loss} depicts that all the models learn efficiently, improve performance via epochs, and no overfitting occurs. 

\begin{figure}[!t]
    \centering
    \includegraphics[width=0.6\textwidth]{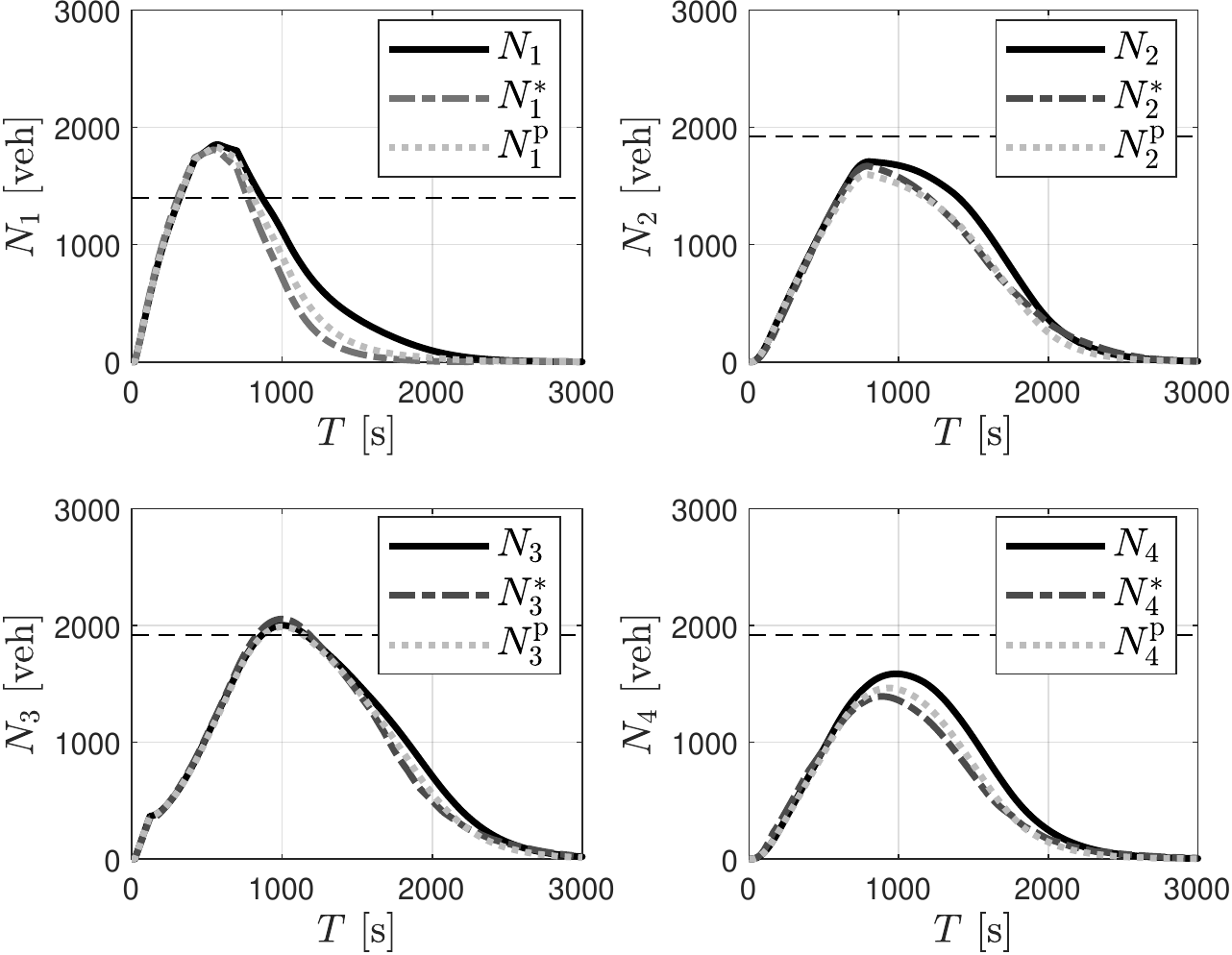}
    \caption{Accumulation trajectories $N_{IJ}$ for aggregated accumulation results QDUE and DSO for $R_1$ -- $R_4$.}
    \label{fig:n_comp_all}
\end{figure}

Thus, the models can be included in the framework and serve as online prediction engines for optimal
generalized costs and are then utilized for the price calculation. The results of the online application result in accumulation trajectories as shown in Figure~\ref{fig:n_comp_all}. Note that the solid black line is the QDUE, the dark grey dashed line constitutes the DSO, and the light grey dashed line the pricing results. Finally, the quantitative analysis is given in Table~\ref{tab:final_comp}. 

Applying (a) the prediction models for the optimal generalized costs and (b) the optimal pricing functions show that the model performance can be increased compared to the QDUE. In $R_1$ the peak of the congestion can not be further mitigated with pricing compared to the DSO. Nevertheless, the congestion dissolves slightly faster than in the QDUE. Afterward, the vehicle accumulation trajectory of the pricing result is slightly higher than the DSO but shows improvement compared to QDUE. As denoted in Table~\ref{tab:final_comp} the DSO improved the situation by 21.93\%; the pricing methodology allows the improvement of TS$_1$ by 14.32\%. In $R_2$ and $R_3$ the pricing methodology even increases the improvement of TS$_2$ and TS$_3$ from the DSO (8.77\% and 4.39\%) to 12.15\% and 5.23\%, respectively. In $R_4$ the improvement of TS$_4$ by applying pricing is 13.47\% compared to 15.00\% when operating in the DSO. Overall, the pricing methodology can decrase the TTS in the network by 10.71\%.
The TTD shows an improvement of 8.62\%. This is surprisingly higher than in the DSO with 7.87\%. The difference in vehicle accumulation $N$ is again 0 supporting the correct formulation of the methodology. 

\begin{table}[!t]
\centering
\caption{Comparison of performance metrices for the QDUE and DSO, respectively. Performance improvements (stated as Impr.) and the difference of vehicles (stated as Diff.) are denoted separately.}
\begin{tabular}{lrrr}
\toprule
  & QDUE   &  Impr./Diff. (DSO) &  Impr./Diff. (Pricing) \\
  & [veh$\cdot$h$\cdot$10$^5$] &  [\%] & [\%] \\
 \midrule
TS$_1$             & 17.97       &  21.93    &  14.32       \\
TS$_2$             &  23.03      &  8.77     &  12.15       \\
TS$_3$            & 27.72        &  4.39     &  5.23  \\
TS$_4$            & 19.42        &  15.00    &  13.47 \\  
TTS          & 88.15          &  11.45    &  10.71 
\\
\midrule
\midrule
 & QDUE   &  DSO &  Impr./Diff. \\
 & [veh$\cdot$km$\cdot$10$^6$] &  [\%] &  [\%] \\
 \midrule
 TTD          &  54.23  &  7.87 &  8.62 \\
\midrule
\midrule
 & QDUE   &  DSO &  Impr./Diff. \\
 & [veh$\cdot$10$^3$] & [veh] & [veh] \\
 \midrule
$N$           & 14.54         &  0   & 0  \\
\bottomrule
\end{tabular}
\label{tab:final_comp}
\end{table}

\begin{figure}[!t]
    \centering
    \includegraphics[width=0.6\textwidth]{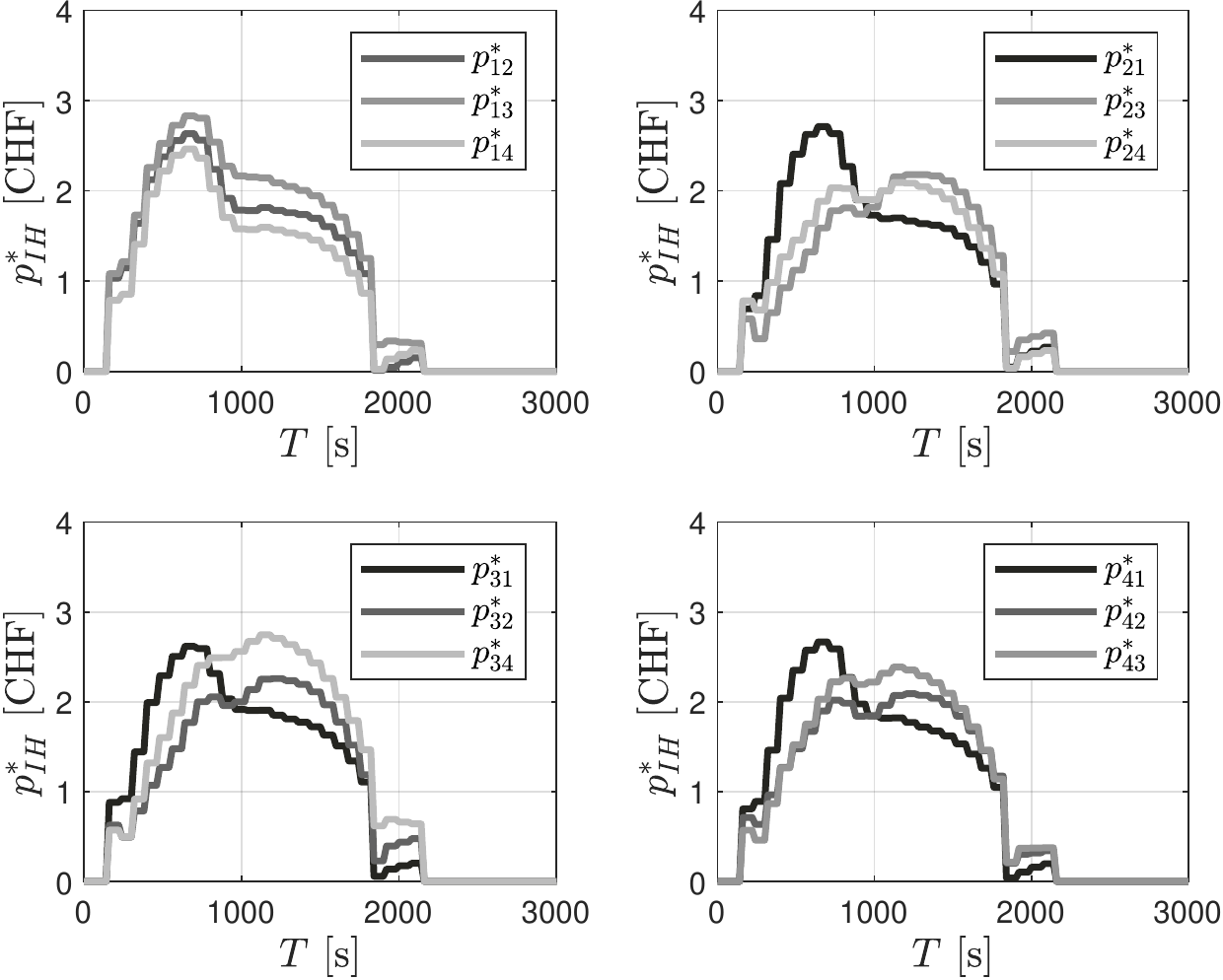}
    \caption{Derived tolls for $R_1$ -- $R_4$; every region is controlled by three tolls $p^*_{IH}$.}
    \label{fig:prices}
\end{figure}

The pricing function for all 12 tolls derived during the online application of our pricing methodology are depicted in Figure~\ref{fig:prices}. Also, we compute average prices for all tolls during the devices are active. The results indicate which toll sets -- on average -- the highest and lowest price (Table~\ref{tab:prices}.). Note that the diagonal elements with $I=H$ are not available, as there are no tolls for trips that do not traverse any region border (marked with NA for not available).  

All the implemented tolls are active 2/3 of the simulation time. For leaving $R_1$ the highest price has to be paid when traversing through $R_3$: The price increases with rising demand in the network but shows the highest peak around 800 seconds of simulation time. Additionally, the prices remain longer above 2 CHF than the prices when traversing through $R_2$ or $R_4$. On average (when the tolls are active) the prices are 1.53 CHF, 1.74 CHF, and 1.35 CHF to traverse through $R_2, R_3, R_4$, respectively (Table~\ref{tab:final_comp}).
In $R_2$ the pricing functions to traverse through the neighbors $R_1, R_3, R_4$ show different behavior. One can note that the $p^*_{21}$ shows the highest price with 2.71 CHF around 800 seconds of simulation time, complying with the experienced congestion in $R_1$, i.e., the dynamic pricing reacts to influence the user's route guidance. Additionally, the alternatives are priced lower, i.e., $p^*_{23} < p^*_{21}$ and $p^*_{24} < p^*_{21}$. After the congestion dissolves, the price to traverse through the city center decreases and remains lower than the other alternatives. The average prices from Table~\ref{tab:final_comp} are computed for $R_2, R_3$ and $R_4$ as follows: 1.46 CHF, 1.35 CHF, and 1.39 CHF. 
Also, tolls that regulate transfer flows starting from $R_3$ to enter the city center ($R_1$) react according to the experienced congestion. Nevertheless, the highest magnitude is computed for traversing to region $R_4$. The averages prices are computed as follows: 1.53 CHF for $p^*_{31}$, 1.43 CHF for $p^*_{32}$, and 1.74 CHF for $p^*{34}$. This also shows that on average, the pricing methodology sets an incentive for user's to traverse through $R_2$ for the first 1000 seconds of the simulation. In $R_4$ again, the highest pricing peak is shown for traversing the city center; after the congestion dissolves, the pricing remains high for traversing through $R_3$. The averages prices are computed with 1.49 CHF, 1.41 CHF, and 1.51 CHF, for $R_1$, $R_2$, and $R_3$, respectively. 

\begin{table}[!t]
\centering
\caption{Average prices for all tolls $p_{IH}$ in [CHF].}
\begin{tabular}{lrrrr}
\toprule
$p^*_{IH}$  [CHF] & $H = 1$      & $H = 2$      & $H = 3$      & $H = 4$      \\
\midrule
$I = 1$ & NA    & 1.53 & 1.74 & 1.35 \\
$I = 2$ & 1.46  & NA   & 1.35  & 1.39 \\
$I = 3$ & 1.53  & 1.43 & NA    & 1.74  \\
$I = 4$ & 1.49  & 1.41  & 1.51 & NA   \\
\bottomrule
\end{tabular}
\label{tab:prices}
\end{table}


\section{CONCLUSION}
\label{sec:conclusion}
The paper presents the derivation of optimal price functions for a multi-region network with homogeneous regions characterized by well-defined MFD functions. First, the optimal routing information (splitting rates) is derived with an LRHO optimization problem, providing network system optimum, which can be utilized as an ideal target for determining dynamic pricing functions. A linearization methodology was implemented to relax the nonlinear optimization problem that allows the application of LRHO. The proposed method from the literature was extended and utilized for obtaining optimal splitting rates in the multi-region network. Accumulation trajectories are utilized to show the system improvement of the methodology with TS for every region, and TTS and TTD for the entire network as performance indicators. The results are compared to the QDUE scenario, which is derived by utilizing the Dijkstra route choice algorithm and an MNL. The proposed linear program reduces TTS significantly and guarantees an optimal and fast solution instead of nonlinear formulations. The computed performance metrics show an improvement in the TTS of 11.45\% and improved TTD of 7.87\%. 

Consequently, an optimal pricing methodology is designed to target pricing network users according to the difference between QDUE and DSO. MNL models are trained for every implemented toll in the network to determine the optimal pricing functions. Thus, the generalized costs of DSO can be predicted accurately. A price matrix is computed online with the utilization of the generalized cost matrices of QDUE and DSO. The online computation and application allow the real-time derivation of prices that reflect the current traffic state in the network and simulate the user's route choice reaction to pricing. The framework allows improving the QDUE state by 10.71\% for the TTS and by 8.62\% for the TTD. Hence, the pricing functions significantly push the multi-region network towards an operation closer to the DSO. 

Future research should focus on running more extensive experiments with different toll set-ups i.e., only specific tolls are active (e.g., tolls protecting the city center or only border region tolls). Also, the prediction of the generalized costs should be further investigated; e.g., the performance of the framework with other route choice model implementations is an interesting research direction. Based on recent research, the simulation plant should also be extended with a trip length model (for now, only average trip lengths are considered), allowing extensive analysis of users' travel times in the system. Furthermore, a weighting of the different regions can be applied in the optimization procedure to account for the different region parameters (i.e., size, storage capacity, etc.). This improves the quality of the modeling further and also contributes to further developments of the proposed methodology. 

\section*{ACKNOWLEDGMENTS}
The authors would like to thank Kimia Chavoshi and Daniel Tschernutter for their valuable discussions about the methodology and technicalities. 

\bibliographystyle{elsarticle-harv}
\bibliography{bibliography}

\newpage

\end{document}